%% file: manuscript-omega-net.tex
\documentclass[final,5p]{elsarticle}

\usepackage[draft]{hyperref}
\usepackage{lineno,amssymb,bm,amsmath,caption,subcaption,booktabs,rotating,tikz,cleveref,gensymb}
\usetikzlibrary{decorations.pathreplacing}
\usepackage[percent]{overpic}
\usepackage[export]{adjustbox}
\usepackage[normalem]{ulem}
\usepackage{dvmacros}
\nolinenumbers
\newcommand\hl[1]{%
  \bgroup
  \hskip0pt\color{black!80!black}%
  #1%
  \egroup
}
\let\linenumbers\nolinenumbers\nolinenumbers
\journal{Medical Image Analysis}

\biboptions{authoryear}
\begin{document}

\input{statistics.tex}

\input{./data/table_2_stats.tex}
\input{./data/model-params.tex}

\begin{frontmatter}

\title{\omeganet{} \hl{(Omega-Net)}: Fully Automatic, Multi-View Cardiac MR Detection, Orientation, and Segmentation with Deep Neural Networks}

\author[oxford,nih,tufts]{Davis M. Vigneault\corref{mycorrespondingauthor}}
\ead{davis.vigneault@gmail.com}

\author[oxford]{Weidi Xie\corref{mycorrespondingauthor}}
\author[brigham]{Carolyn Y. Ho}
\author[wisc]{David A. Bluemke}
\author[oxford]{J. Alison Noble}

\address[oxford]{Institute of Biomedical Engineering, Department of Engineering, University of Oxford}
\address[nih]{Department of Radiology and Imaging Sciences, Clinical Center, National Institutes of Health}
\address[tufts]{Tufts University School of Medicine, Sackler School of Graduate Biomedical Sciences}
\address[brigham]{Cardiovascular Division, Brigham and Women's Hospital}
\address[wisc]{University of Wisconsin-Madison, School of Medicine and Public Health}

\cortext[mycorrespondingauthor]{These authors contributed equally to this work.}

\begin{abstract}

\input{./text/abstract}

\end{abstract}

\begin{keyword}
cardiac magnetic resonance \sep semantic segmentation \sep deep convolutional neural networks \sep spatial transformer networks
\MSC[2010] 00-01\sep  99-00
\end{keyword}

\end{frontmatter}

\linenumbers

\newcommand{\figdir}{./figures/}
\newcommand{\tabdir}{./tables/}

%%%%%%%%%%%%%%%%%%%%%%%%%%%%%%%%%%%%%%%%%%%%%%%%%%%%%%%%%%%%%%%%%%%%%%%%%%%%%%%%
%% Introduction
%%%%%%%%%%%%%%%%%%%%%%%%%%%%%%%%%%%%%%%%%%%%%%%%%%%%%%%%%%%%%%%%%%%%%%%%%%%%%%%%

\section{Introduction} \label{introduction}

\input{./text/introduction.tex}

%%%%%%%%%%%%%%%%%%%%%%%%%%%%%%%%%%%%%%%%%%%%%%%%%%%%%%%%%%%%%%%%%%%%%%%%%%%%%%%%
%% Methods
%%%%%%%%%%%%%%%%%%%%%%%%%%%%%%%%%%%%%%%%%%%%%%%%%%%%%%%%%%%%%%%%%%%%%%%%%%%%%%%%

\section{Methods} \label{methods}

\input{./text/methods.tex}

%%%%%%%%%%%%%%%%%%%%%%%%%%%%%%%%%%%%%%%%%%%%%%%%%%%%%%%%%%%%%%%%%%%%%%%%%%%%%%%%
%% Experiments
%%%%%%%%%%%%%%%%%%%%%%%%%%%%%%%%%%%%%%%%%%%%%%%%%%%%%%%%%%%%%%%%%%%%%%%%%%%%%%%%

\section{Experiments} \label{experiments}

\input{./text/experiments.tex}

%%%%%%%%%%%%%%%%%%%%%%%%%%%%%%%%%%%%%%%%%%%%%%%%%%%%%%%%%%%%%%%%%%%%%%%%%%%%%%%%
%% Results
%%%%%%%%%%%%%%%%%%%%%%%%%%%%%%%%%%%%%%%%%%%%%%%%%%%%%%%%%%%%%%%%%%%%%%%%%%%%%%%%

\section{Results} \label{results}

\input{./text/results.tex}

%%%%%%%%%%%%%%%%%%%%%%%%%%%%%%%%%%%%%%%%%%%%%%%%%%%%%%%%%%%%%%%%%%%%%%%%%%%%%%%%
%% Discussion
%%%%%%%%%%%%%%%%%%%%%%%%%%%%%%%%%%%%%%%%%%%%%%%%%%%%%%%%%%%%%%%%%%%%%%%%%%%%%%%%

\section{Discussion} \label{discussion}

\input{./text/discussion.tex}

%%%%%%%%%%%%%%%%%%%%%%%%%%%%%%%%%%%%%%%%%%%%%%%%%%%%%%%%%%%%%%%%%%%%%%%%%%%%%%%%
%% Summary
%%%%%%%%%%%%%%%%%%%%%%%%%%%%%%%%%%%%%%%%%%%%%%%%%%%%%%%%%%%%%%%%%%%%%%%%%%%%%%%%

\section{Summary} \label{summary}

\input{./text/summary.tex}

%%%%%%%%%%%%%%%%%%%%%%%%%%%%%%%%%%%%%%%%%%%%%%%%%%%%%%%%%%%%%%%%%%%%%%%%%%%%%%%%
%% Acknowledgements
%%%%%%%%%%%%%%%%%%%%%%%%%%%%%%%%%%%%%%%%%%%%%%%%%%%%%%%%%%%%%%%%%%%%%%%%%%%%%%%%

\section*{Acknowledgements} \label{acknowledgements}

\input{./text/acknowledgements.tex}

%%%%%%%%%%%%%%%%%%%%%%%%%%%%%%%%%%%%%%%%%%%%%%%%%%%%%%%%%%%%%%%%%%%%%%%%%%%%%%%%
%% References
%%%%%%%%%%%%%%%%%%%%%%%%%%%%%%%%%%%%%%%%%%%%%%%%%%%%%%%%%%%%%%%%%%%%%%%%%%%%%%%%

\section*{References} \label{references}

\bibliography{citations}

\end{document}

%% file: statistics.tex
%%%%%%%%%%
% MICCAI %
%%%%%%%%%%

% Jaccard

\newcommand{\ACDCONJLVBP}{0.912}
\newcommand{\ACDCONJRVBP}{0.852}
\newcommand{\ACDCONJLVMY}{0.803}

\newcommand{\ACDCFINJLVBP}{0.896}
\newcommand{\ACDCFINJRVBP}{0.832}
\newcommand{\ACDCFINJLVMY}{0.826}

\newcommand{\ACDCFIOJLVBP}{0.869}
\newcommand{\ACDCFIOJRVBP}{0.784}
\newcommand{\ACDCFIOJLVMY}{0.775}

% Dice

\newcommand{\ACDCONDLVBP}{0.954}
\newcommand{\ACDCONDRVBP}{0.920}
\newcommand{\ACDCONDLVMY}{0.891}

\newcommand{\ACDCFINDLVBP}{0.945}
\newcommand{\ACDCFINDRVBP}{0.908}
\newcommand{\ACDCFINDLVMY}{0.905}

\newcommand{\ACDCFIODLVBP}{0.930}
\newcommand{\ACDCFIODRVBP}{0.879}
\newcommand{\ACDCFIODLVMY}{0.873}

% Data Set

\newcommand{\SpacingMU}{1.3}
\newcommand{\SpacingSD}{0.2}
\newcommand{\MatrixMU}{253.53}
\newcommand{\MatrixSD}{46.73}

\newcommand{\NumPtC}{21}
\newcommand{\NumPtO}{42}
\newcommand{\NumPtT}{63}
\newcommand{\N}{256}
\newcommand{\NumFramesMin}{20}
\newcommand{\NumFramesMax}{50}

% Cross Validation
\newcommand{\NumFolds}{3}
\newcommand{\NumImFoldA}{4477}
\newcommand{\NumImFoldB}{4750}
\newcommand{\NumImFoldC}{4625}

% Architecture
\newcommand{\networkmomentum}{0.9}
\newcommand{\weightdecay}{10^{-4}}
%\newcommand{\trainablekernels}{128}

% Augmentation
\newcommand{\AugTrans}{0.15}
\newcommand{\AugRot}{\hl{15}}
\newcommand{\AugZoom}{0.15}

\newcommand{\AUC}{0.992}

\newcommand{\SAXAlvendCGlobalMu}{-24.9}
\newcommand{\SAXAlvendCGlobalSd}{7.8}
\newcommand{\SAXAlvendOGlobalMu}{-29.8}
\newcommand{\SAXAlvendOGlobalSd}{13.6}
\newcommand{\SAXAlvendpGlobal}{NS}
\newcommand{\SAXAlvepiCGlobalMu}{-10.3}
\newcommand{\SAXAlvepiCGlobalSd}{3.5}
\newcommand{\SAXAlvepiOGlobalMu}{-11.8}
\newcommand{\SAXAlvepiOGlobalSd}{13.7}
\newcommand{\SAXAlvepipGlobal}{NS}
\newcommand{\SAXBlvendCGlobalMu}{-28.0}
\newcommand{\SAXBlvendCGlobalSd}{4.1}
\newcommand{\SAXBlvendOGlobalMu}{-28.6}
\newcommand{\SAXBlvendOGlobalSd}{6.3}
\newcommand{\SAXBlvendpGlobal}{NS}
\newcommand{\SAXBlvepiCGlobalMu}{-13.2}
\newcommand{\SAXBlvepiCGlobalSd}{3.1}
\newcommand{\SAXBlvepiOGlobalMu}{-10.6}
\newcommand{\SAXBlvepiOGlobalSd}{3.5}
\newcommand{\SAXBlvepipGlobal}{<0.01}
\newcommand{\SAXElvendCGlobalMu}{-24.7}
\newcommand{\SAXElvendCGlobalSd}{4.4}
\newcommand{\SAXElvendOGlobalMu}{-28.7}
\newcommand{\SAXElvendOGlobalSd}{6.0}
\newcommand{\SAXElvendpGlobal}{<0.05}
\newcommand{\SAXElvepiCGlobalMu}{-8.9}
\newcommand{\SAXElvepiCGlobalSd}{2.7}
\newcommand{\SAXElvepiOGlobalMu}{-7.7}
\newcommand{\SAXElvepiOGlobalSd}{3.2}
\newcommand{\SAXElvepipGlobal}{NS}

%% file: data/table_2_stats.tex
\newcommand{\AALZe}{0.783}
\newcommand{\AAMZe}{0.834}
\newcommand{\AAHZe}{0.871}
\newcommand{\ASLZe}{0.789}
\newcommand{\ASMZe}{0.843}
\newcommand{\ASHZe}{0.880}
\newcommand{\AHLZe}{0.765}
\newcommand{\AHMZe}{0.819}
\newcommand{\AHHZe}{0.855}
\newcommand{\AVLZe}{0.788}
\newcommand{\AVMZe}{0.831}
\newcommand{\AVHZe}{0.863}
\newcommand{\BALZe}{0.793}
\newcommand{\BAMZe}{0.841}
\newcommand{\BAHZe}{0.876}
\newcommand{\BSLZe}{0.800}
\newcommand{\BSMZe}{0.848}
\newcommand{\BSHZe}{0.884}
\newcommand{\BHLZe}{0.780}
\newcommand{\BHMZe}{0.831}
\newcommand{\BHHZe}{0.861}
\newcommand{\BVLZe}{0.787}
\newcommand{\BVMZe}{0.832}
\newcommand{\BVHZe}{0.864}
\newcommand{\BALOn}{0.819}
\newcommand{\BAMOn}{0.857}
\newcommand{\BAHOn}{0.885}
\newcommand{\BSLOn}{0.820}
\newcommand{\BSMOn}{0.862}
\newcommand{\BSHOn}{0.891}
\newcommand{\BHLOn}{0.812}
\newcommand{\BHMOn}{0.845}
\newcommand{\BHHOn}{0.871}
\newcommand{\BVLOn}{0.822}
\newcommand{\BVMOn}{0.856}
\newcommand{\BVHOn}{0.882}
\newcommand{\CALZe}{0.792}
\newcommand{\CAMZe}{0.841}
\newcommand{\CAHZe}{0.875}
\newcommand{\CSLZe}{0.797}
\newcommand{\CSMZe}{0.849}
\newcommand{\CSHZe}{0.883}
\newcommand{\CHLZe}{0.779}
\newcommand{\CHMZe}{0.830}
\newcommand{\CHHZe}{0.861}
\newcommand{\CVLZe}{0.793}
\newcommand{\CVMZe}{0.830}
\newcommand{\CVHZe}{0.863}
\newcommand{\CALOn}{0.816}
\newcommand{\CAMOn}{0.856}
\newcommand{\CAHOn}{0.884}
\newcommand{\CSLOn}{0.820}
\newcommand{\CSMOn}{0.862}
\newcommand{\CSHOn}{0.890}
\newcommand{\CHLOn}{0.804}
\newcommand{\CHMOn}{0.843}
\newcommand{\CHHOn}{0.869}
\newcommand{\CVLOn}{0.818}
\newcommand{\CVMOn}{0.855}
\newcommand{\CVHOn}{0.883}
\newcommand{\CALTw}{0.816}
\newcommand{\CAMTw}{0.857}
\newcommand{\CAHTw}{0.885}
\newcommand{\CSLTw}{0.819}
\newcommand{\CSMTw}{0.862}
\newcommand{\CSHTw}{0.890}
\newcommand{\CHLTw}{0.805}
\newcommand{\CHMTw}{0.844}
\newcommand{\CHHTw}{0.869}
\newcommand{\CVLTw}{0.819}
\newcommand{\CVMTw}{0.857}
\newcommand{\CVHTw}{0.884}
\newcommand{\DALZe}{0.797}
\newcommand{\DAMZe}{0.844}
\newcommand{\DAHZe}{0.877}
\newcommand{\DSLZe}{0.798}
\newcommand{\DSMZe}{0.851}
\newcommand{\DSHZe}{0.886}
\newcommand{\DHLZe}{0.781}
\newcommand{\DHMZe}{0.832}
\newcommand{\DHHZe}{0.861}
\newcommand{\DVLZe}{0.804}
\newcommand{\DVMZe}{0.842}
\newcommand{\DVHZe}{0.869}
\newcommand{\DALOn}{0.819}
\newcommand{\DAMOn}{0.856}
\newcommand{\DAHOn}{0.884}
\newcommand{\DSLOn}{0.821}
\newcommand{\DSMOn}{0.862}
\newcommand{\DSHOn}{0.890}
\newcommand{\DHLOn}{0.805}
\newcommand{\DHMOn}{0.839}
\newcommand{\DHHOn}{0.868}
\newcommand{\DVLOn}{0.828}
\newcommand{\DVMOn}{0.858}
\newcommand{\DVHOn}{0.884}
\newcommand{\DALTw}{0.821}
\newcommand{\DAMTw}{0.858}
\newcommand{\DAHTw}{0.886}
\newcommand{\DSLTw}{0.822}
\newcommand{\DSMTw}{0.863}
\newcommand{\DSHTw}{0.892}
\newcommand{\DHLTw}{0.811}
\newcommand{\DHMTw}{0.843}
\newcommand{\DHHTw}{0.870}
\newcommand{\DVLTw}{0.831}
\newcommand{\DVMTw}{0.860}
\newcommand{\DVHTw}{0.886}
\newcommand{\DALTh}{0.821}
\newcommand{\DAMTh}{0.858}
\newcommand{\DAHTh}{0.886}
\newcommand{\DSLTh}{0.822}
\newcommand{\DSMTh}{0.863}
\newcommand{\DSHTh}{0.892}
\newcommand{\DHLTh}{0.811}
\newcommand{\DHMTh}{0.843}
\newcommand{\DHHTh}{0.869}
\newcommand{\DVLTh}{0.830}
\newcommand{\DVMTh}{0.859}
\newcommand{\DVHTh}{0.886}

%% file: data/model-params.tex
\newcommand{\MillionsOfParamsA}{7.0}
\newcommand{\MillionsOfParamsB}{3.5}
\newcommand{\MillionsOfParamsC}{4.5}
\newcommand{\MillionsOfParamsD}{5.5}

%% file: text/abstract.tex
Pixelwise segmentation of the left ventricular (\LV{}) myocardium and the four cardiac chambers in \ND{2} steady state free precession (\SSFP{}) cine sequences is an essential preprocessing step for a wide range of analyses.
Variability in contrast, appearance, orientation, and placement of the heart between patients, clinical views, scanners, and protocols makes fully automatic semantic segmentation a notoriously difficult problem.
Here, we present \omeganet{} (Omega-Net): a novel convolutional neural network (\CNN{}) architecture for simultaneous \hl{localization}, transformation into a canonical orientation, and semantic segmentation.
First, an \hl{initial} segmentation is performed on the input image; second, the features learned during this \hl{initial} segmentation are used to predict the parameters needed to transform the input image into a canonical orientation; and third, a \hl{final} segmentation is performed on the transformed image.
In this work, \omeganet{}s of varying depths were trained to detect five foreground classes in any of three clinical views (short axis, \SA{}; four-chamber, \HLA{}; two-chamber, \VLA{}), without prior knowledge of the view being segmented.
This constitutes a substantially more challenging problem compared with prior work.
The architecture was trained on a cohort of patients with hypertrophic cardiomyopathy (\HCM{}, $N = \NumPtO{}$) and healthy control subjects ($N = \NumPtC{}$).
Network performance\hl{,} as measured by weighted foreground intersection-over-union (\IoU{})\hl{,} was substantially improved for the best-performing \omeganet{} compared with \UNet{} segmentation without \hl{localization} or orientation ($0.858$ vs $0.834$).
\hl{
In addition, to be comparable with other works, \omeganet{} \hl{was retrained} from scratch on the publicly available 2017 MICCAI Automated Cardiac Diagnosis Challenge (\miccaidata{}) dataset.
The \omeganet{} outperformed the state-of-the-art method in segmentation of the \LV{} and \RV{} bloodpools, and performed slightly worse in segmentation of the \LV{} myocardium.
We conclude that} this architecture represents a substantive advancement over prior approaches, with implications for biomedical image segmentation more generally.

%% file: text/introduction.tex
\input{\figdir omega-net-architecture.tex}

Pixelwise segmentation of the left ventricular (\LV{}) myocardium and the four cardiac chambers in \ND{2} steady state free precession (\SSFP{}) cine sequences is an essential preprocessing step for volume estimation (e.g., ejection fraction, stroke volume, and cardiac output); morphological characterization (e.g., myocardial mass, regional wall thickness and thickening, and eccentricity); and strain analysis \citep{Peng2016}.
However, automatic cardiac segmentation remains a notoriously difficult problem, given:

\begin{itemize}
\item Biological variability in heart size, orientation in the thorax, and morphology (both in healthy subjects and in the context of disease).
\item Variability in contrast and image appearance with different scanners, protocols, and clinical planes.
\item Interference of endocardial trabeculation and papillary muscles.
\item Poorly defined borders between the ventricles and the atria, as well as between the chambers and the vasculature.
\end{itemize}

Three broad approaches have been employed to address this complexity.
First, the scope of the problem can be restricted, i.e., to segmentation of the \LV{} myocardium and bloodpool in the \SA{} view only.
Second, user interaction can be used to provide a sensible initialization, supply anatomical landmarks, or correct errors.
Third, prior knowledge of cardiac anatomy may be incorporated into model-based approaches.  
Clearly, none of these approaches is ideal: the first limiting the information which can be gleaned from the algorithm; the second being labor-intensive for the clinician; and the third requiring careful construction of algorithmic constraints.

Recently, deep convolutional neural networks (\CNN{}s) have been \hl{proposed} to great effect both in \hl{natural} image classification \citep{Krizhevsky2012,Simonyan2015}, and segmentation \citep{Long2015,Noh2015,Yu2016}, \hl{as well as for} biomedical image analysis \citep{Ronneberger2015,Xie2015}.
\CNN{} segmentation of short axis \CMR{} has been applied to the \LV{} blood-pool \citep{Tan2016,Poudel2016a,Tan2017}, the \RV{} blood-pool \citep{Luo2016}, and both simultaneously \citep{Tran2016,Lieman-Sifry2017,Vigneault2017}.
In each of these methods, either \hl{localization} and segmentation \hl{were} performed separately \citep{Tan2016, Poudel2016a, Tan2017, Luo2016}, or \hl{the images were manually cropped such that the heart was in the image center and took up a majority of the image, obviating the \hl{localization} task} \hl{\citep{Tran2016, Lieman-Sifry2017, Vigneault2017}}.
Neither end-to-end \hl{localization} and segmentation nor transformation into a canonical orientation prior to segmentation has been described.

\input{\figdir canonical-orientation.tex}

\hl{
In the Deep Learning~(DL) literature, \CNN{}s were only designed to be invariant to small perturbations by average/max pooling.
However, in essense, the square-windowed convolution~(correlation) operations have several limitations, e.g., they are neither rotation invariant nor equivariant, nor scale invariant, and therefore require large datasets representing all possible rotations and/or substanial data augmentations~\citep{Sifre2013, Dieleman2015}.
}
In this paper, we propose the \omeganet{} (Omega-Net), a novel \CNN{} architecture trained end-to-end to tackle three important tasks: \hl{localization}, transformation into a canonical orientation, and segmentation (Fig.~\ref{fig:omega-net-architecture}).

For simplicity, we use the \UNet{} as the fundamental component of the \hl{initial} and \hl{final} segmentation modules \citep{Ronneberger2015}, though more advanced networks such as ResNet \citep{He2016} could be substituted in\hl{stead}.
Inspired by the spatial transformer network \citep{Jaderberg2015}, we designed a fully differentiable architecture that \hl{simultaneously} achieves \hl{localization} and transformation into a canonical orientation.

The transformed image is then fed into a \hl{final} segmentation module, which resembles the stacked hourglass architecture \citep{Newell2016}.
\hl{
In a stacked hourglass, segmentation is performed by stacking two or more \UNet{}-like modules in series, where the features learned by one \UNet{} serve as the input to its successor, and intermediate segmentations are predicted at the output of each \UNet{}.
This architecture has been shown to produce progressively more accurate predictions, with diminishing returns at each stage \citep{Newell2016}.
}

We demonstrate that the \omeganet{} is capable of the fully automatic segmentation of five foreground classes (\LV{} myocardium, the left and right atria, and the left and right ventricles) in three orthogonal clinical planes (short axis, \SA{}; four-chamber, \HLA{}; and two-chamber, \VLA{}), with simultaneous rigid transformation of the input into a canonical orientation (defined separately for each view, Fig.~\ref{fig:canonical-orientation}).
Moreover, the network is trained on a multicenter population \hl{\citep{Ho2017}} of patients with hypertrophic cardiomyopathy (\HCM{}), which increases the complexity of the problem due to the highly variable appearance of the \LV{} in these patients.
\hl{N}etwork performance as measured by weighted foreground intersection-over-union (\IoU{}) was substantially improved in the best-performing \omeganet{} 
compared with \UNet{} segmentation without \hl{localization} and orientation alignment ($0.858$ vs $0.834$).
\hl{
In addition, we retrained the network from scratch on the 2017 MICCAI Automated Cardiac Diagnosis Challenge (\miccaidata{}) dataset,\footnote{\url{https://www.creatis.insa-lyon.fr/Challenge/acdc/}} and achieved results which outperform the current state-of-the-art \citep{Isensee2018} in terms of \LV{} and \RV{} cavity segmentation, and perform slightly worse in terms of \LV{} myocardium segmentation.
}

%% file: figures/omega-net-architecture.tex
\renewcommand{\captiontitle}{Overview of the \omeganet{} architecture}
\begin{figure*}
\begin{center}

\begin{tikzpicture}
\newcommand{\tw}{\textwidth}

\node [anchor=south west] (image) at (0,0) {\includegraphics[clip, trim=0 260 0 260, width=1.0\textwidth]{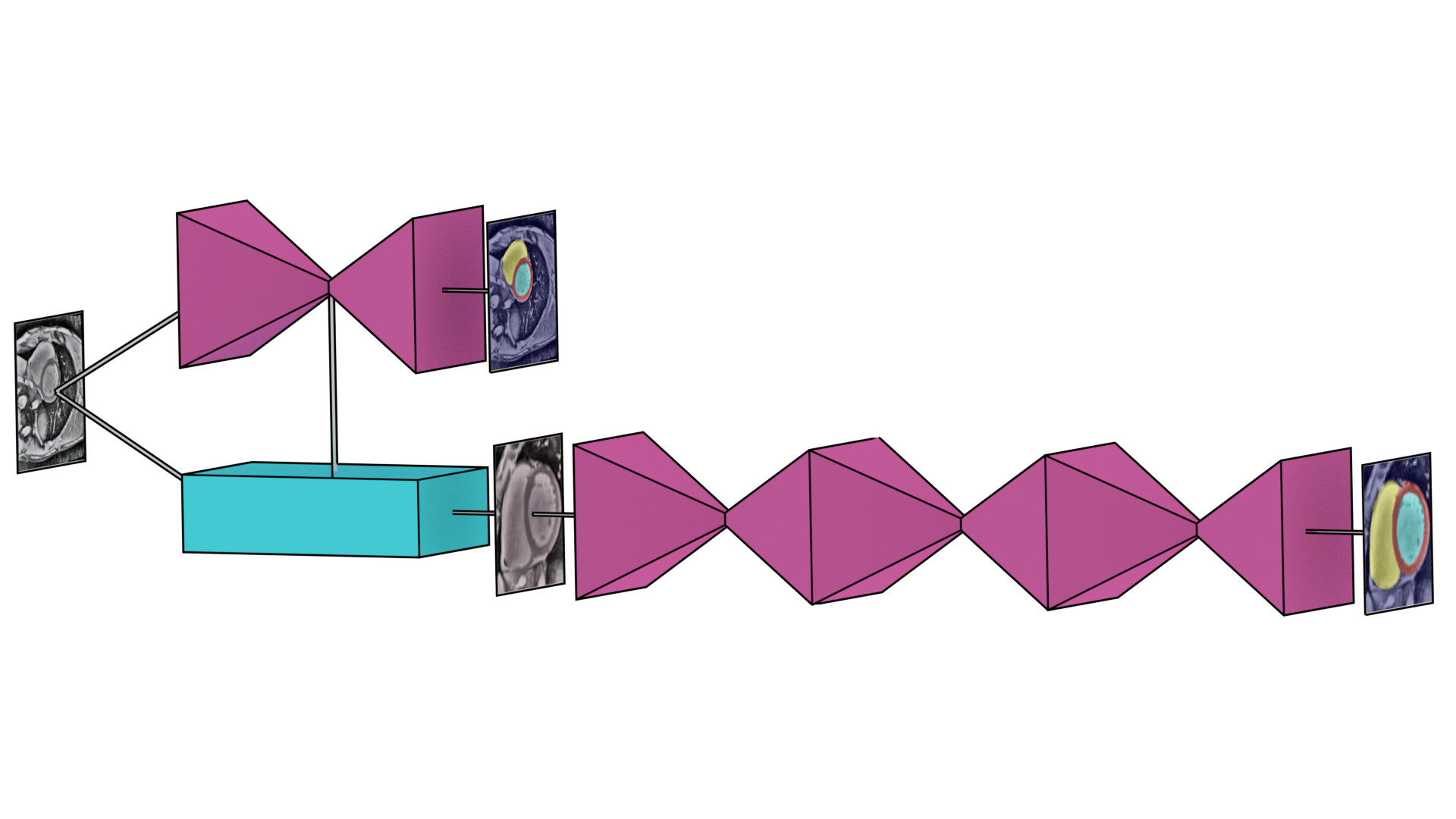}};

\node [anchor=south west] (a) at (0.025\tw,0.22\tw) {\large $\image$};
\node [anchor=south west] (b) at (0.35\tw,0.13\tw) {\large $\image^\prime$};
\node [anchor=south west] (c) at (0.35\tw,0.29\tw) {\large $S$};
\node [anchor=south west] (d) at (0.95\tw,0.13\tw) {\large $S^\prime$};
\node [anchor=south west] (e) at (0.17\tw,0.06\tw) {\large $\trans(\image, \tmat)$};

\draw [decorate,decoration={brace,amplitude=10pt},xshift=0pt,yshift=0pt](0.1275\tw,0.3\tw) -- (0.34\tw,0.3\tw) node [black,midway,xshift=0.0cm,yshift=+0.7cm] {(a) \hl{Initial} Segmentation Module};
\draw [decorate,decoration={brace,amplitude=10pt},xshift=0pt,yshift=0pt](0.3425\tw,0.045\tw) -- (0.1325\tw,0.045\tw) node [black,midway,xshift=0.0cm,yshift=-0.7cm] {(b) \hl{Transformation} Module};
\draw [decorate,decoration={brace,amplitude=10pt},xshift=0pt,yshift=0pt](0.4\tw,0.15\tw) -- (0.935\tw,0.15\tw) node [black,midway,xshift=0.0cm,yshift=+0.7cm] {(c) \hl{Final} Segmentation Module};
\end{tikzpicture}

\caption[\captiontitle]{\captiontitle{}.  (a) The initial, unoriented \SSFP{} image $\image$ is fed into a \UNet{} module, producing an \hl{initial} segmentation $S$.  (b) The features from the central (most downsampled) layers of this \UNet{} are used by the \hl{transformation} module to predict the parameters $\tmat$ of a rigid, affine transformation and transform the input image into a cannonical orientation, $\image^\prime = \trans(\image, \tmat)$.  (c) This transformed image is fed into a stacked hourglass module to obtain a \hl{final} segmentation in the canonical orientation $S^\prime$. Note that, all modules shown are trained in an end-to-end way from scratch.}
\label{fig:omega-net-architecture}
\end{center}
\end{figure*}

%% file: figures/canonical-orientation.tex
\renewcommand{\captiontitle}{Orthogonal clinical views in canonical orientation}
\begin{figure*}
\begin{center}

\setlength{\tabcolsep}{1pt}

\begin{tabular}{ccccc}

\toprule
\SA{} (Base) & \SA{} (Midslice) & \SA{} (Apex) & \HLA{} & \VLA{} \\
\midrule

\multicolumn{5}{c}{Original Untransformed Image: $\image$} \\

\includegraphics[width=0.19\textwidth]{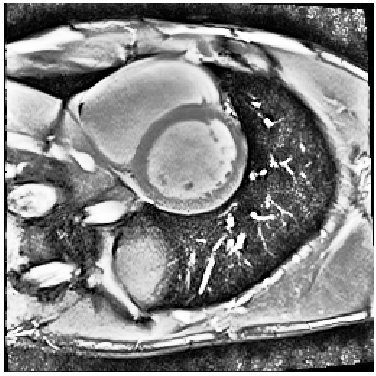} &
\includegraphics[width=0.19\textwidth]{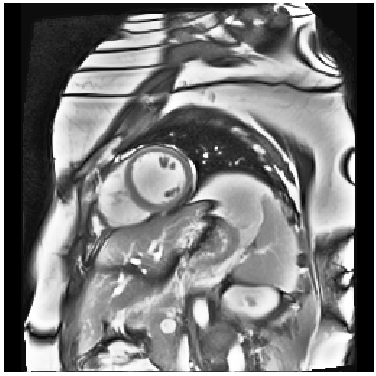} &
\includegraphics[width=0.19\textwidth]{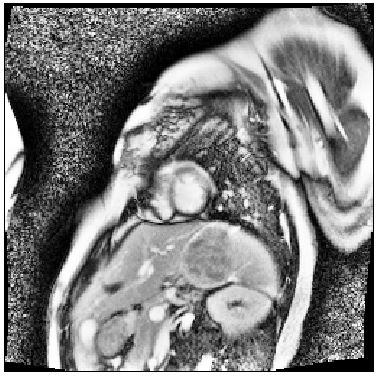} &
\includegraphics[width=0.19\textwidth]{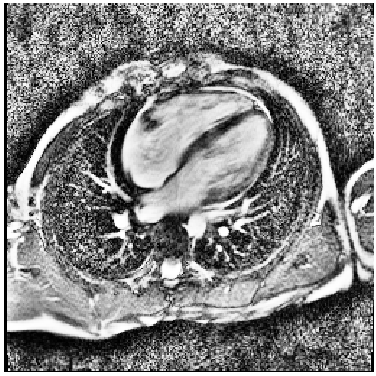} &
\includegraphics[width=0.19\textwidth]{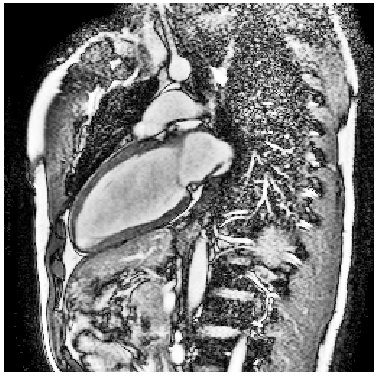} \\

\multicolumn{5}{c}{Structure \hl{Localization}} \\

\includegraphics[width=0.19\textwidth]{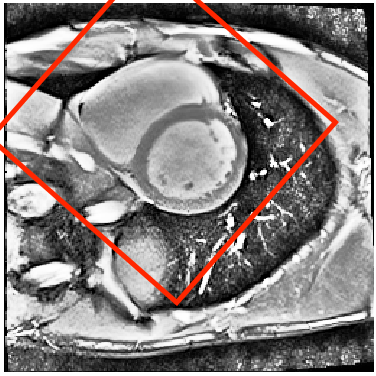} &
\includegraphics[width=0.19\textwidth]{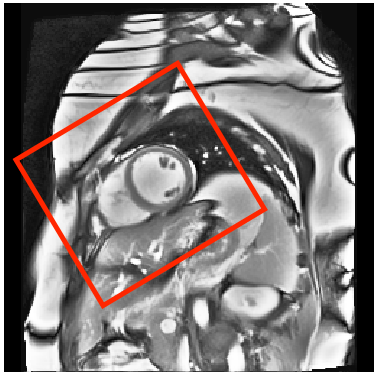} &
\includegraphics[width=0.19\textwidth]{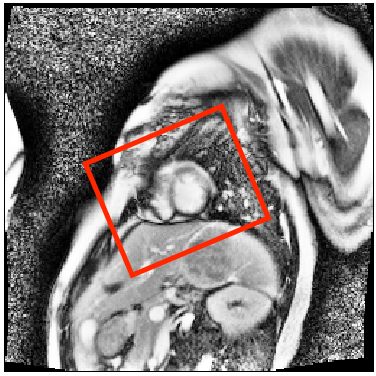} &
\includegraphics[width=0.19\textwidth]{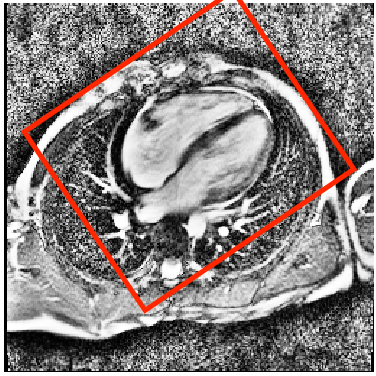} &
\includegraphics[width=0.19\textwidth]{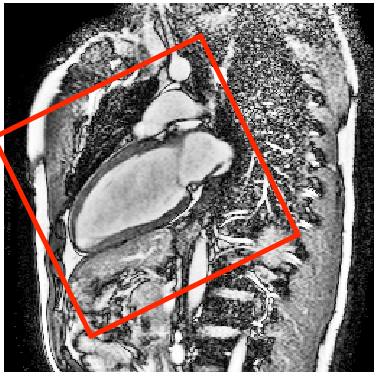} \\

\multicolumn{5}{c}{Image in Canonical Orientation: $\image^\prime = \trans(\image,M)$} \\

\includegraphics[width=0.19\textwidth]{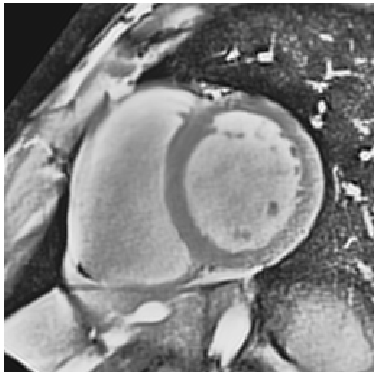} &
\includegraphics[width=0.19\textwidth]{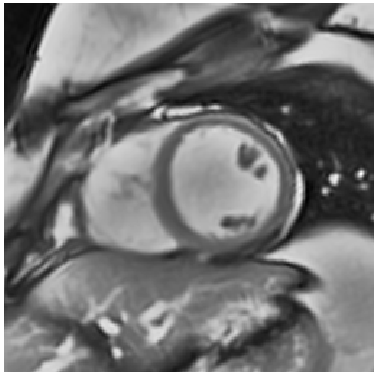} &
\includegraphics[width=0.19\textwidth]{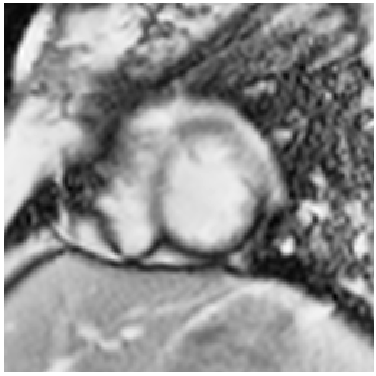} &
\includegraphics[width=0.19\textwidth]{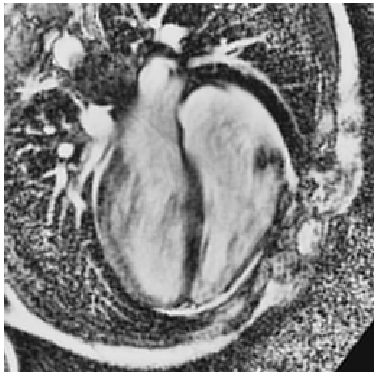} &
\includegraphics[width=0.19\textwidth]{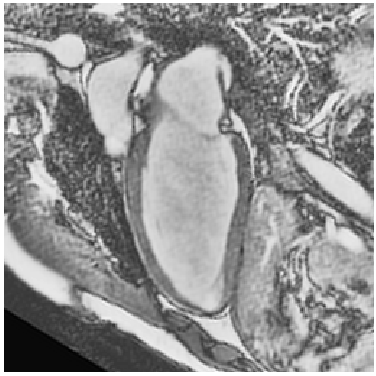} \\
\bottomrule

\end{tabular}

\caption[\captiontitle]{\captiontitle{}.
Representative short axis (\SA{}), four-chamber (\HLA{}), and two-chamber (\VLA{}) images are shown as acquired (top), and having undergone rigid, affine transformation into a canonical orientation (bottom).
Consistent with common clinical practice, the heart is rotated such that in the \SA{} views, the right ventricle appears on the (radiological) right side of the image, whereas in the \HLA{} and \VLA{} views, the long axis of the left ventricle is oriented vertically.
The heart is also centered and scaled to fill $90\%$ of the image.
Note the heterogeneity in size, orientation, and appearance of the heart in the untransformed images, which contributes to the difficulty of segmentation.
}
\label{fig:canonical-orientation}
\end{center}
\end{figure*}

%% file: text/methods.tex
\hl{
Due to the lack of rotation invariance/equivariance in \CNN{}s, current practice is for models to be trained with large datasets representing all possible rotations and/or substantial data augmentations (e.g., affine transformations, warpings, etc).
We conjecture that b}iomedical image segmentation can be more efficiently accomplished if structures of interest have first been detected and transformed into a canonical orientation.
In the context of \CMR{}, the canonical orientation is defined separately for each clinical plane (Fig.~\ref{fig:canonical-orientation}).
We propose a \hl{stepwise} strategy for segmentation of cardiac \SSFP{} images in an end-to-end differentiable \CNN{} framework, allowing for the \hl{localization}, alignment, and segmentation tasks to be codependent. 
Our model consists of three stages.
First, the full-resolution, \hl{original} input image $\image$ undergoes \hl{an initial} segmentation using a \UNet{} module (\S\ref{sec:unet}).
Second, the central (most \hl{down-sampled}) features of the aforementioned \UNet{} module are used to predict a rigid, affine matrix $\tmat$ capable of transforming $\image$ into a canonical orientation $\timage = \trans(\image, \tmat)$ (\S\ref{sec:stn}).
Third, the transformed image $\timage$ is segmented using a stacked hourglass module (\S\ref{sec:hourglass}).
In the following subsections, each component of the network is discussed in detail.
In terms of notation, a superposed chevron (e.g., $\hat{x}$) indicates ground truth, and a superscript tick (e.g., $x^\prime$) indicates that the quantity pertains to the transformed data.

\subsection{Initial segmentation (\UNet{}) module}\label{sec:unet}

\input{\figdir unet-module.tex}

The proposed network makes use of the \UNet{} module (Fig.~\ref{fig:unet-module}), a type of deep convolutional neural network which has performed well in biomedical segmentation tasks \citep{Long2015,Ronneberger2015,Xie2015}.
The \UNet{} architecture consists of a down-sampling path (left) followed by an up-sampling path (right) to restore the original spatial resolution.
The downsampling path resembles the canonical classification \CNN{} \citep{Krizhevsky2012,Simonyan2015}, with two $3 \times 3$ convolutions, a rectified linear unit (\ReLU{}) activation, 
and a $2 \times 2$ max pooling step repeatedly applied to the input image and feature maps.  
In the upsampling path, the reduction in spatial resolution is ``undone'' by performing $2 \times 2$ up-sampling, \ReLU{} activation, and $3 \times 3$ convolution, eventually mapping the intermediate feature representation back to the original resolution.
To provide accurate boundary localization, skip connections are used, where feature representations from the down-sampling path are concatenated with feature maps of the same resolution in the up-sampling path.
Batch normalization \mbox{\citep{Ioffe2015}}, which has been shown to counteract gradient vanishing and to lead to better convergence, was performed between each pair of convolution and ReLU activation layers.
The loss $L_{S_U}$ for the \UNet{} module is the categorical cross entropy between the output of the softmax layer, $P$, and the ground truth segmentation, $\hat{S}$,

\begin{equation}\label{eqn:unet-loss}
L_{S_U} = -\frac{1}{HW} \sum_{\forall h,w} \CCE(P_{h,w}, \hat{S}_{h,w}),
\end{equation}

\noindent where

\begin{equation}\label{eqn:cross-entropy}
\CCE(x, \hat{x}) = - \hat{x} \log(x) + (1 - \hat{x}) \log(1 - x).
\end{equation}

\hl{
Here, $H$ and $W$ are the height and width of the input image in pixels, and $h$ and $w$ are corresponding pixel indices.
}

\subsection{\hl{Transformation} module}\label{sec:stn}

\input{\figdir stn-module.tex}

The spatial transformer network (\STN{}) was originally proposed as a general layer for classification tasks requiring spatial invariance for high performance \hl{\citep{Jaderberg2015}}.
The \STN{} module itself consists of three submodules, namely: a localization network (\LocNet{}), which predicts a rigid, affine transformation matrix, $\tmat$; a grid generator, which implements the transform, $\trans$; and a sampler, which implements the interpolation.

In \citep{Jaderberg2015}, 
the \STN{} was allowed to learn whichever transformation parameters best aid the classification task; 
no ground truth transformation was specified, 
and the predicted transformation matrix was used to transform the intermediate \emph{feature maps}.
By contrast, in our application we are specifically interested in learning to transform the \emph{input image} into the standard clinical orientation, as a precursor to semantic segmentation.

%%%%%%%%%%%%%%%%%%%%%%%%%%%%%%%%%%%%%%%%%%
% 	Locnet
%%%%%%%%%%%%%%%%%%%%%%%%%%%%%%%%%%%%%%%%%%
\subsubsection{Localization network (\LocNet{})}

Intuitively, a human expert is able to provide translation, rotation, and scaling information given a rough segmentation of the heart.  Based on this assumption, we branch out a small localization network (\LocNet{}) from the layer immediately following the final max pooling step of the \UNet{} in order to predict the transformation parameters (Fig.~\ref{fig:stn-module}).  As we have restricted our transform to \hl{allow only translation, rotation, and scaling}, the affine matrix was decomposed into three separate matrices:

$$
\tmat = SRT,
$$

\noindent where $T$ is the translation matrix:

$$
T =
\begin{bmatrix}
1 & 0 & t_x \\
0 & 1 & t_y \\
0 & 0 & 1 \\
\end{bmatrix};
$$

\noindent $R$ is the (counter clockwise) rotation matrix:

$$
R =
\begin{bmatrix}
\cos(\theta) & - \sin(\theta) & 0 \\
\sin(\theta) & \cos(\theta) & 0 \\
0 & 0 & 1 \\
\end{bmatrix};
$$

\noindent and $S$ is the (uniform) scaling matrix:

$$
S =
\begin{bmatrix}
s & 0 & 0 \\
0 & s & 0 \\
0 & 0 & 1 \\
\end{bmatrix}.
$$

\noindent Note that the images are defined on a normalized coordinate space $\{x, y\} \in [-1, +1]$, such that rotation and scaling occur relative to the image center.

In practice, the \LocNet{} learns to predict only the relevant parameters, $\tparams = \begin{bmatrix}t_x & t_y & \theta & s \end{bmatrix}^\top$.
During training, we explicitly provide the ground-truth transformation parameters $\hat{\tparams} = \begin{bmatrix} \hat{t_x} & \hat{t_y} & \hat{\theta} & \hat{s} \end{bmatrix}$, minimizing two types of losses, which we term \emph{matrix losses} and \emph{image losses}.

The matrix losses are regression losses between the ground truth and predicted parameters ($L_{t_x}$, $L_{t_y}$, $L_\theta$, $L_s$).
For scaling and translation, mean squared error (\MSE{}) was used:

\begin{align}
L_{t_x} & = \frac{1}{2} (t_x - \hat{t_x})^2, \label{eqn:attn-mat-loss-tx} \\
L_{t_y} & = \frac{1}{2} (t_y - \hat{t_y})^2, \mathrm{~and} \label{eqn:attn-mat-loss-ty} \\
L_s     & = \frac{1}{2} (s - \hat{s})^2. \label{eqn:attn-mat-loss-s}
\end{align}

\input{\figdir wrapped-phase-loss.tex}

Na\"{i}ve \MSE{} is an inappropriate loss for regressing on $\theta$ given its periodicity\hl{.  Intuitively, t}his can be understood intuitively by considering ground truth and predicted rotations of $\hat{\theta} = +\pi$ and $\theta = -\pi$, which yield a high \MSE{} in spite of being synonymous.
For this reason, we introduce a wrapped phase loss, mean squared wrapped error (\MSWE{}, Fig.~\ref{fig:wrapped-phase-loss}), where $\theta - \hat{\theta}$ is wrapped into the range $[-\pi, \pi)$ prior to calculating the standard \MSE{},

\begin{equation}
L_\theta = \frac{1}{2} \left(\wrap(\theta - \hat{\theta})\right)^2, \label{eqn:attn-mat-loss-r}
\end{equation}

\noindent and the wrapping operator $\wrap$ is defined as

$$
\wrap(\cdot) = \mod(\cdot + \pi, 2\pi) - \pi.
$$

Training the \hl{transformation} module based on these losses alone caused the network to overfit the training data.
For this reason, we additionally regularized based on the \MSE{} between the input image after translation, rotation, and scaling with the ground truth ($\hat{\tparams}$) and predicted ($\tparams$) transformation parameters:

\begin{align}
L_{\image_t}      & = \frac{1}{2} (\trans(\image, T  ) - \trans(\image, \hat{T}  ))^2,                     \label{eqn:attn-image-loss-t} \\
L_{\image_\theta} & = \frac{1}{2} (\trans(\image, RT ) - \trans(\image, \hat{R}\hat{T} ))^2, \mathrm{~and} \label{eqn:attn-image-loss-r} \\
L_{\image_s}      & = \frac{1}{2} (\trans(\image, SRT) - \trans(\image, \hat{S}\hat{R}\hat{T}))^2.         \label{eqn:attn-image-loss-s}
\end{align}

%%%%%%%%%%%%%%%%%%%%%%%%%%%%%%%%%%%%%%%%%%
% 	Sampler
%%%%%%%%%%%%%%%%%%%%%%%%%%%%%%%%%%%%%%%%%%
\subsubsection{Grid generation and sampling}

\hl{In general,} a \ND{2} ``grid generator'' takes a (typically uniform) sampling of points $G \in \Real^{2 \times H' \times W'}$ and transforms them according to the parameters predicted by a \LocNet{}.
In our application, we created three such grids, each of equal dimension to the input ($H' = W' = \N$) and uniformly spaced over the extent of the image ($x \in [-1,1]$, $y \in [-1, 1]$)\hl{.
T}hese grids were then transformed by the matrices $T$, $RT$, and $SRT$ (predicted by the \LocNet{}) to determine which points to sample from the input image.

The ``sampler'' takes the input image $\image \in \Real^{H \times W \times C}$ and the transformed grid $G^\prime$ as arguments, and produces a resampled image $\timage \in \Real^{H' \times W' \times C}$.\footnote{\hl{For completeness, we have included the number of channels $C$ in this description as a variable parameter; however, it should be emphasized that in our application the grayscale input image $\image$ is transformed, such that $C = 1$.}}
For each channel $c \in [1 \twodots C ]$, the output $\timage_{h',w',c}$ at the location $(h',w')$ is a weighted sum of the input values $\image_{h,w,c}$ in the neighborhood of location ($G^\prime_{1,h',w'}, G^\prime_{2,h',w'}$),

$$
\begin{aligned}
\timage_{h',w',c} & = \sum^H_{h=1}\sum^W_{w=1}\image_{h, w, c} \\
                  & \cdot \max\left(0, 1-|\alpha_v G^\prime_{1, h', w'} + \beta_v - h| \right) \\
                  & \cdot \max\left(0, 1-|\alpha_u G^\prime_{2, h', w'} + \beta_u - w| \right),
\end{aligned}
$$

\noindent where

$$
\begin{aligned}
\alpha_v & = +\frac{H-1}{2},               \\
\beta_v  & = -\frac{H+1}{2},              \\
\alpha_u & = +\frac{W-1}{2}, \mathrm{ and} \\
\beta_u  & = - \frac{W+1}{2}.
\end{aligned}
$$

\noindent Every step here is differentiable (either a gradient or sub-gradient is defined), such that the model can be trained end-to-end.

%%%%%%%%%%%%%%%%%%%%%%%%%%%%%%%%%%%%%%%%%%
% 	Hourglass
%%%%%%%%%%%%%%%%%%%%%%%%%%%%%%%%%%%%%%%%%%
\subsection{\hl{Final} segmentation (stacked hourglass) module}\label{sec:hourglass}

The output of the \hl{transformation} module, having been transformed into a canonical orientation, is then input into a stacked hourglass architecture.
The hourglass consisted of $D = [1 \twodots 3]$ \UNet{} modules in series with one another, each producing a segmentation $S_{H,d}$, where $d \in [1 \twodots D]$.
With reference to Eqn.~\eqref{eqn:cross-entropy}, the categorical cross-entropy between the softmax output of the hourglass at depth $d$, $P^{H,d}_{h,w}$ and the (transformed) ground truth $\hat{S}^\prime$ segmentations is calculated,

\begin{equation}\label{eqn:hourglass-loss}
L_{S_{H,d}} = -\frac{1}{HW} \sum_{\forall h,w} \CCE(P^{H,d}_{h,w}\hat{S}^\prime_{h,w}).
\end{equation}

%%%%%%%%%%%%%%%%%%%%%%%%%%%%%%%%%%%%%%%%%%
% 	Summary
%%%%%%%%%%%%%%%%%%%%%%%%%%%%%%%%%%%%%%%%%%
\subsubsection{Summary}

To summarize, we train the \omeganet{} with one loss from the \hl{initial} segmentation module, \cref{eqn:unet-loss}; four matrix losses, \cref{eqn:attn-mat-loss-tx,eqn:attn-mat-loss-ty,eqn:attn-mat-loss-s,eqn:attn-mat-loss-r}, and three image losses, \cref{eqn:attn-image-loss-t,eqn:attn-image-loss-s,eqn:attn-image-loss-r}, from the \hl{transformation} module; and between one and three losses from the \hl{final} segmentation module, \cref{eqn:hourglass-loss}.  Therefore, the overall loss function may be written:

\begin{equation}
\begin{aligned}
L_\Omega & = \alpha_1 L_{S_U} \\
  & + \alpha_2 (L_{t_x} + L_{t_y} + L_{\theta}+ L_{s}) \\
  & + \alpha_3 (L_{I_t} + L_{I_\theta} + L_{I_s}) \\
  & + \alpha_4 \sum_{d=1}^D L_{S_{H,d}}, \\
\end{aligned}
\end{equation}

\noindent where $\alpha_1 = 100.0$, $\alpha_2 = 100.0$, $\alpha_3 = 0.1$, and $\alpha_4 = 1.0$.
The architectures \hl{evaluated} are summarized in Table~\ref{tab:architecture-descriptions}.

\hl{
While the dataset was manually augmented by transforming the input with small, rigid, affine transformations, it is worth noting that data augmentation is performed \emph{implicitly} in the fine segmentation module by virtue of the fact that, in the early stages of training, the transformation parameters predicted by the transformation module are random.
}

%% file: figures/unet-module.tex
\renewcommand{\captiontitle}{\UNet{} module}
\begin{figure}
\centering
\includegraphics[clip, trim=0 0 0 0,width=0.5\textwidth]{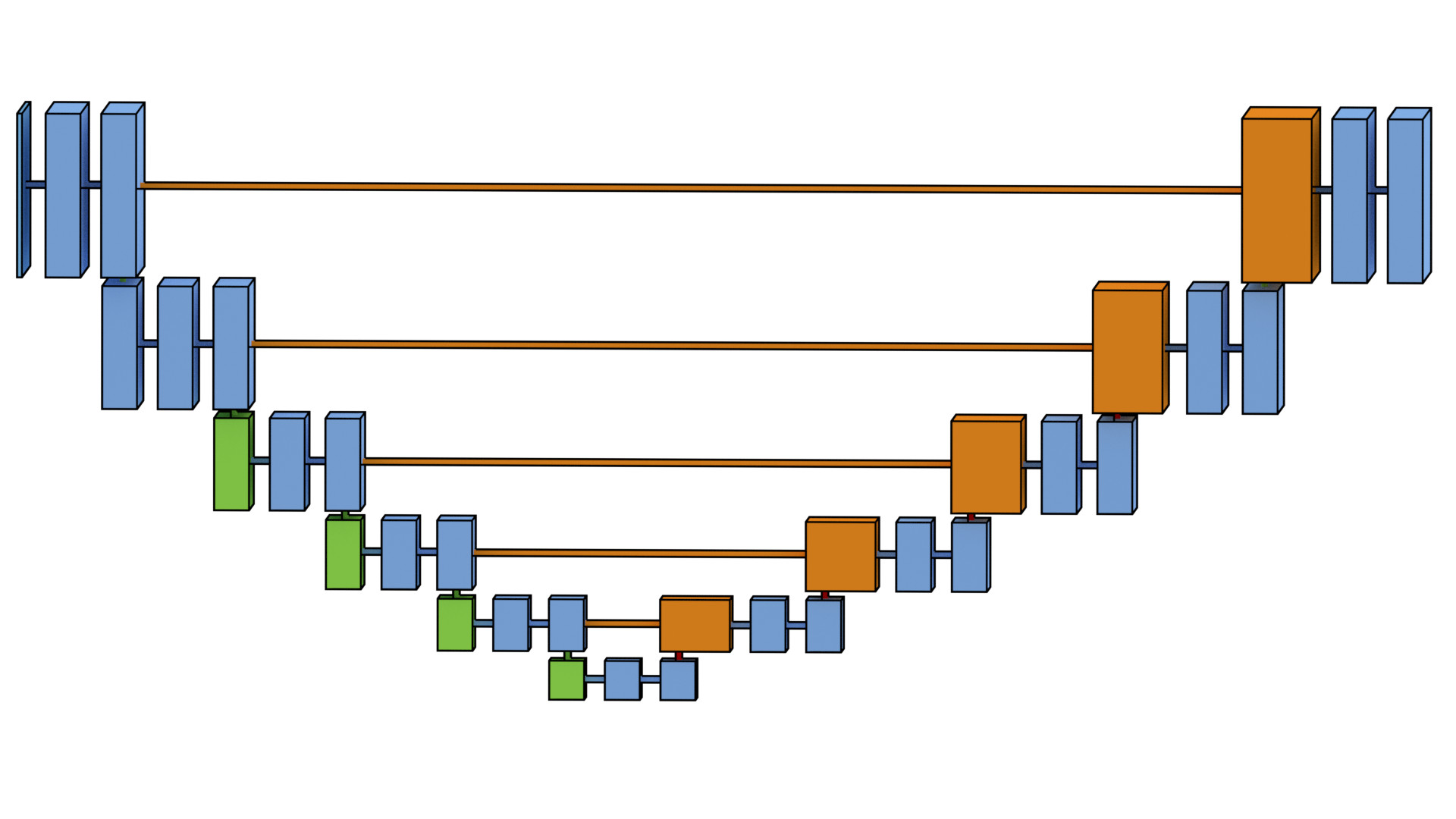}
\caption[\captiontitle]{\captiontitle{}.  The input image is of size $\N \times \N \times N$, where $N$ is the number of channels ($1$ for all networks).  
\hl{Blue, green, and orange boxes correspond to multichannel feature maps.  Green indicates a downsampled feature map.  Orange indicates the result of a copy, concatenated with an upsampled feature map}.}
\label{fig:unet-module}
\end{figure}

%% file: figures/stn-module.tex
\renewcommand{\captiontitle}{Spatial transformer network module}
\begin{figure}
\begin{center}

\begin{tikzpicture}
\node [anchor=south west] (image) at (0,0) {\includegraphics[width=0.5\textwidth]{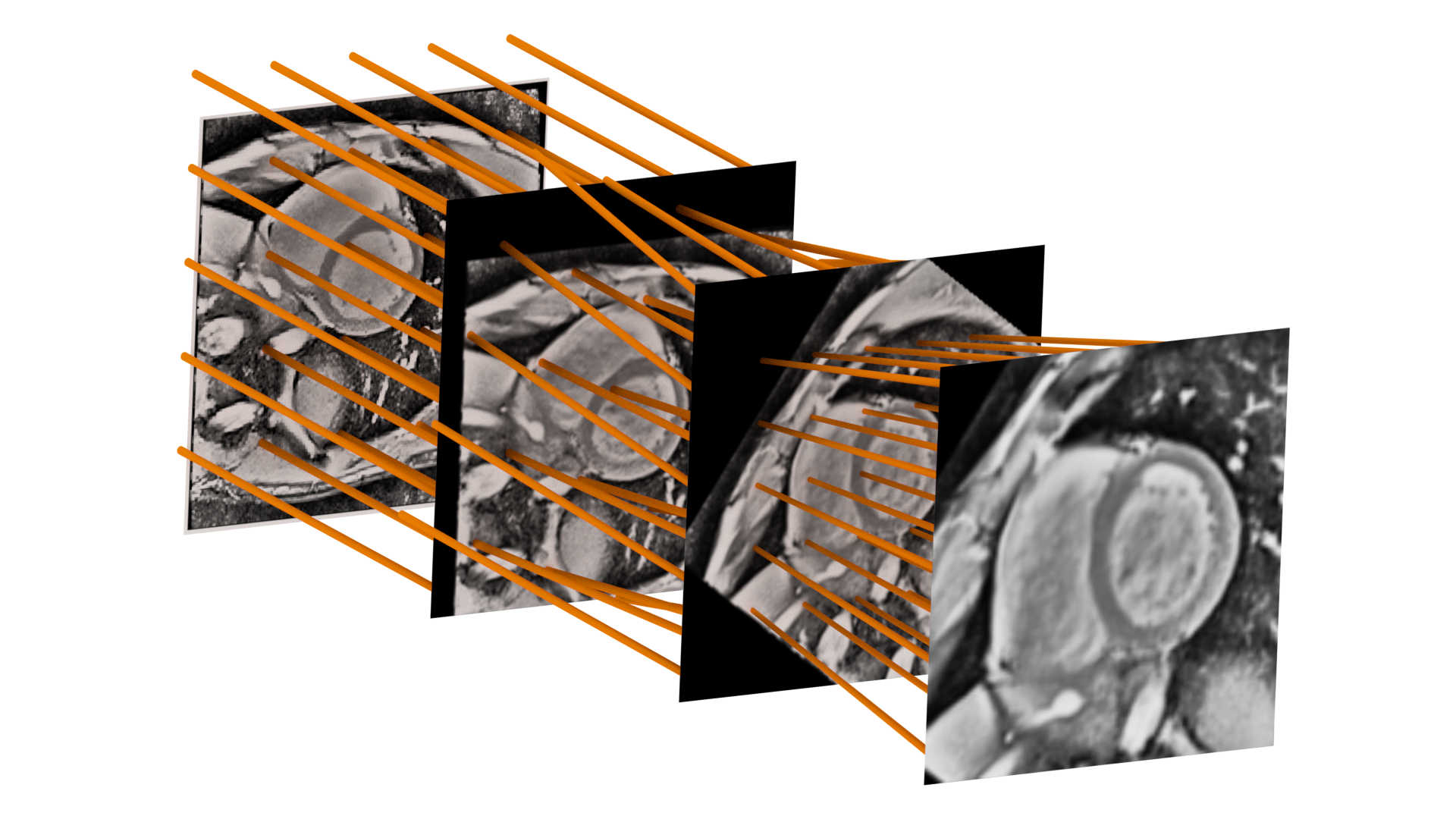}};
\draw [decorate,decoration={brace,amplitude=10pt},xshift=0pt,yshift=0pt](2.8,1.317) -- (1.25,1.85) node [black,midway,xshift=-0.3cm,yshift=-0.7cm] {\footnotesize $\image_2 = \trans(\image, T)$};
\draw [decorate,decoration={brace,amplitude=10pt},xshift=0pt,yshift=0pt](4.35,0.783) -- (2.8,1.317) node [black,midway,xshift=-0.3cm,yshift=-0.7cm] {\footnotesize $\image_3 = \trans(\image_2, R)$};
\draw [decorate,decoration={brace,amplitude=10pt},xshift=0pt,yshift=0pt](5.9,0.25) -- (4.35,0.783) node [black,midway,xshift=-0.3cm,yshift=-0.7cm] {\footnotesize $\trans(\image_3, S)$};
\end{tikzpicture}

\caption[\captiontitle]{\captiontitle{}.  Note that in the actual implementation, all transformations are performed relative to the input image $\image$ (i.e., $\trans(\image, T)$, $\trans(\image, RT)$, and $\trans(\image, SRT)$); for clarity, the transformations have been presented here as successive steps.}
\label{fig:stn-module}
\end{center}
\end{figure}
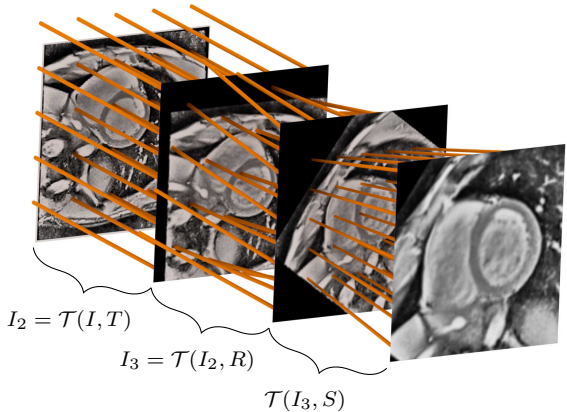

%% file: figures/wrapped-phase-loss.tex
\renewcommand{\captiontitle}{Mean squared error (\MSE{}) vs mean squared wrapped error (\MSWE{})}
\begin{figure*}
\begin{center}
\begin{subfigure}{0.48\textwidth}
\includegraphics[clip, trim=90 30 50 70,width=1.0\textwidth]{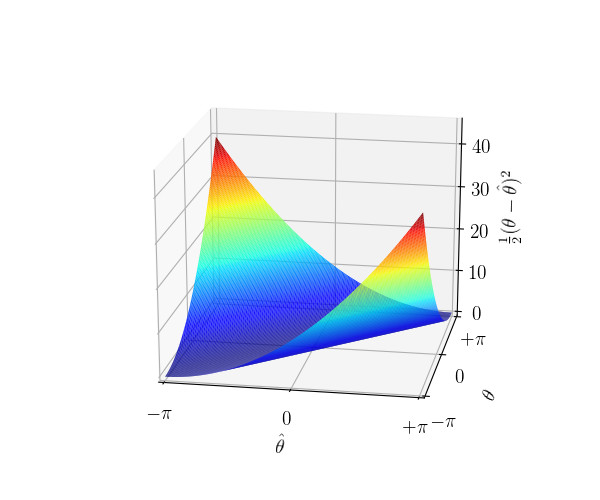}
\caption{Mean Squared Error (\MSE{})}
\end{subfigure}
\begin{subfigure}{0.48\textwidth}
\includegraphics[clip, trim=90 30 50 70,width=1.0\textwidth]{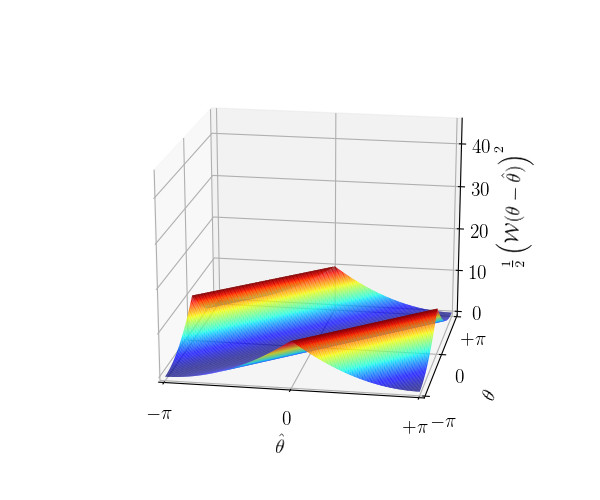}
\caption{Mean Squared Wrapped Error (\MSWE{})}
\end{subfigure}
\caption[\captiontitle]{\captiontitle{}.}
\label{fig:wrapped-phase-loss}
\end{center}
\end{figure*}

%% file: text/experiments.tex
\subsection{\hl{HCMNet dataset}}

\input{\tabdir architecture-descriptions.tex}

The \hl{HCMNet dataset} consisted of \NumPtT{} subjects: \NumPtO{} patients with overt hypertrophic cardiomyopathy (\HCM{}) and \NumPtC{} healthy control subjects \hl{\citep{Ho2017}}.
\CMR{} was performed with a standardized protocol at 10 \hl{clinical sites} from 2009 to 2011.
Nine centers used 1.5-T magnets, and one used a 3-T magnet.
Where available, three \SA{} (basal, equatorial, and apical), one \HLA{}, and one \VLA{} \SSFP{} cine series were obtained.
\hl{
Images had an in-plane spacing of $\SpacingMU{} \pm \SpacingSD{}$mm and matrix size of $\MatrixMU{} \pm \MatrixSD{}$ pixels; further details concerning the \CMR{} acquisition are given in the supplement to \citet{Ho2017}.
}

The \LV{} myocardium, and all four cardiac chambers were manually segmented the \SA{}, \HLA{}, and \VLA{} views (noting that not all classes are visible in the \SA{} and \VLA{} views).
\hl{
\ND{2}+time volumes were loaded into ITK-Snap \citep{Yushkevich2006}; every fifth frame was segmented manually, and the remaining frames were automatically interpolated.
(Segmentation was performed by the first author, with five years experience in manual \CMR{} segmentation).
}
The papillary muscles and the trabeculation of the \LV{} and \RV{} were excluded from the myocardium.

Each volume was cropped or padded as appropriate to $\N \times \N$ pixels in the spatial dimensions, and varied from $\NumFramesMin{}$ to $\NumFramesMax{}$ frames in the time dimension.
Nonuniform background illumination was corrected by dividing by an estimated illumination field, and background corrected images were histogram equalized.
Each individual image was normalized to zero mean and unit standard deviation before being input into the \CNN{}.

\subsubsection{Training and cross-validation}

For cross-validation, the subjects were partitioned into three folds of approximately equal size ($\NumImFoldA{}$, $\NumImFoldB{}$, and $\NumImFoldC{}$ images, respectively) such that the images from any one subject were present in one fold only.
Each of the four architectures (Table~\ref{tab:architecture-descriptions}) were trained on all three combinations of two folds and tested on the remaining fold.
\hl{
Network A was the \hl{initial} segmentation module alone; since the \UNet{} was performed well in biomedical image segmentation tasks, this was regarded as a strong baseline.
Networks B, C, and D were \omeganet{} architectures with 1, 2, and 3, \UNet{} components in the \hl{final} segmentation module.
}

The networks were initialized with orthogonal weights \hl{\citep{Saxe2013}}, and were optimized using Adam \hl{optimization} \citep{Kingma2015} by minimizing categorical cross-entropy.
The learning rate was initialized to $0.001$ and decayed by $0.1$ every $26$ epochs.
To avoid over-fitting, data augmentation (translations and scaling $\pm \AugTrans{}\%$ of the image width; rotations $\pm \AugRot{}$\degree) and a weight decay of $\weightdecay{}$ was \hl{applied to the input to the initial segmentation module.
Notably, data augmentation is performed \emph{implicitly} in the final segmentation module, due to the fact that the predicted transformation parameters are random early in training.
Note also that data augmentation was performed independently for each time frame.
}

\subsubsection{Measure of performance}

Weighted foreground intersection-over-union (\IoU{}) was calculated \hl{image-by-image} between the prediction and manual segmentations.
For a binary image (one foreground class, one background class), \IoU{} (also known as the Jaccard index) is defined for the ground truth and predicted images $I_T$ and $I_P$ as

\begin{equation}
\IoU{} \left( I_T, I_P \right) = \frac{|I_T \cap I_P|}{|I_T \cup I_P|},
\end{equation}

\noindent noting that a small positive number should be added to the denominator in a practical implementation to avoid division by zero.
To extend this concept to multiclass segmentation, \IoU{} was calculated separately for each foreground class.
A weighted sum of these five \IoU{} values was then calculated, where the weights were given by the ratio between the relevant foreground class and the union of all foreground classes, yielding weighted, mean foreground IoU{}.

\subsubsection{Implementation}

The model was implemented in the Python programming language using the Keras interface to Tensorflow \hl{\citep{Chollet2015,Abadi2016}}, and trained on one NVIDIA Titan X graphics processing unit (GPU) with 12 GB of memory.
For all network architectures, it took roughly 20 minutes to iterate over the entire training set (1 epoch).
At test time, the network predicted segmentations at roughly 15 frames per second.

\hl{
\subsection{2017 MICCAI \miccaidata{} dataset}

Network B \hl{was retrained} from scratch on the 2017 MICCAI \miccaidata{} dataset.
Th\hl{is} training dataset consist\hl{s} of stacked \SA{} cines from 100 patients with a range of pathologies (20 normal, 20 with myocardial infarction, 20 with dilated cardiomyopathy, 20 with hypertrophic cardiomyopathy, and 20 with \RV{} disease).
Ground truth \LV{} myocardium, \LV{} bloodpool, and \RV{} bloodpool segmentations were provided at \ED{} and \ES{} for all spatial slices.
Segmentation performance was assessed using both \IoU{} and the Dice coefficient in order to facilitate comparison with the \miccaidata{} results:

\begin{equation}
\mathrm{Dice} \left( I_T, I_P \right) = \frac{2|I_T \cap I_P|}{|I_T| + |I_P|},
\end{equation}

\noindent The network was trained using five-fold cross-validation, in accordance with the current state-of-the-art \citep{Isensee2017,Isensee2018}.
}

%% file: tables/architecture-descriptions.tex
\renewcommand{\captiontitle}{\CNN{} architecture variants considered}
\begin{table*}
\begin{center}
\begin{tabular}{lrrrrr} \hline
\toprule
Name      & \UNet{} 0 & \UNet{} 1 & \UNet{} 2 & \UNet{} 3 & Millions of Parameters \\
\midrule
Network A & 128       & --        & --        & --        & \MillionsOfParamsA{} \\
Network B & 64        & 64        & --        & --        & \MillionsOfParamsB{} \\
Network C & 64        & 64        & 64        & --        & \MillionsOfParamsC{} \\
Network D & 64        & 64        & 64        & 64        & \MillionsOfParamsD{} \\
\bottomrule
\end{tabular}
\caption[\captiontitle{}]{
\captiontitle{}.
\hl{
Network A is the baseline \UNet{} (the \hl{initial} segmentation module alone, without transformation or \hl{final} segmentation modules).
Networks B, C, and D are full \omeganet{} architectures with 1, 2, and 3 \UNet{} components, respectively, in the fine-grained segmentation module.
\UNet{} 0 is the \UNet{} in the \hl{initial} segmentation module; \UNet{}s 1, 2, and 3 are the first, second, and third \UNet{} components in the \hl{final} segmentation module, as applicable.
For each \UNet{} component of each network variant, the length of the feature vector is provided.
}
}
\label{tab:architecture-descriptions}
\end{center}
\end{table*}

%% file: text/results.tex
\subsection{\hl{HCMNet dataset}}

\input{\tabdir table_2.tex}

\subsubsection{\hl{Segmentation}}

Weighted foreground \IoU{} was calculated separately for each image, and the median and interquartile range (\IQR{}) of all predictions is reported.
As accuracy is not necessarily the same across all clinical planes, the performance of the four networks relative to manual segmentation is reported for all views combined, and also for each clinical plane separately (Table~\ref{tab:architectureaccuracy}).

\input{\figdir hourglass-accuracy.tex}

It is instructive to examine intermediate network performance at each successive \UNet{} (Fig.~\ref{fig:hourglass-accuracy}).

\begin{itemize}
\item Although Network A contains the most parameters, adding the \hl{final} segmentation module \hl{was found to increase} network performance \emph{at the level of the \hl{initial} \UNet{}} compared with Network A; 
i.e., the performance of the \hl{initial} segmentation module \UNet{} (\UNet{} $0$) is $\approx~0.007$ higher in Networks B and C compared with Network A, 
and $\approx~0.003$ higher in Network D compared with Networks B and C.
\item There \hl{wa}s a substantial increase in performance between the \hl{initial} and \hl{final} segmentation \UNet{}s, i.e., \UNet{}s 0 and 1 ($\approx~0.016$, $\approx~0.015$, and $\approx~0.012$ increases for Networks B, C, and D, respectively) .
\item In Networks C and D, there \hl{wa}s not a substantial increase in performance between successive \UNet{}s in the \hl{final} segmentation module.
\end{itemize}

\input{\figdir histograms.tex}

As performance is likely to differ between structures, image\hl{-}wise histograms of foreground \IoU{} are plotted for the best performing network (\bestnetwork{}) for each structure and clinical plane (Fig.~\ref{fig:histograms}).
In all three clinical planes, performance is worst for the \LV{} myocardium, best \hl{for} the \LV{} blood pool, and intermediate \hl{for} the remaining structures.
\hl{Intuitively, r}elatively poor \LV{} myocardial segmentation performance can be understood by considering that segmentation error is concentrated primarily at the structure boundaries.
Therefore, structures with a high ratio of perimeter-to-area (such as the \LV{} myocardium, which has both an internal and external perimeter, i.e., endocardium and epicardium) are predisposed to perform poorly.
A number of factors may contribute to the superior performance of \LV{} bloodpool segmentation.

\begin{itemize}
\item The \LV{} myocardium provides a high-contrast boundary along much of the perimeter of the \LV{} bloodpool.
\item Compared with other cardiac chambers, the \LV{} bloodpool has relatively less anatomical variation between subjects.
\item The three orthogonal planes examined in this study are all defined relative to the left ventricle; therefore, the appearance \hl{of} the \LV{} is more consistent between subjects.
\end{itemize}

\hl{
Fig.~\ref{fig:roc} presents the precision-recall curve, showing the ``success rate'' (vertical axis) defined as the fraction of cases in which weighted foreground \IoU{} exceeded a varying threshold varying from $0.4$ to $1.0$ (horizontal axis).
The resulting precision-recall curve had an area under the curve (AUC) of \AUC{}, demonstrating the accuracy of the \omeganet{}.
The ``failure rate'' can also be calculated from this curve as $1 - \mathrm{success rate}$.
For example, for a conservative definition of failure as weighted foreground \IoU{} $< 0.9$, the failure rate is approximately $1\%$.
}

Representative segmentations produced by \bestnetwork{} in all views are shown for healthy control subjects in Fig.~\ref{fig:representative-results-control} and for patients with overt \HCM{} in Fig.~\ref{fig:representative-results-overt}.
Note that the ground truth segmentations have been transformed by the predicted parameters rather than the ground truth parameters in order to aid interpretation \hl{in these figures}.
The network successfully transformed the images into the canonical orientation for all cases shown.
Notably, the myocardial segmentation consistently excludes papillary muscles and myocardial trabeculation.
Moreover, the network appears to reliably identify the atrioventricular valve plane in the long axis views, which is a useful result deserving of attention in future work.

\input{\figdir roc.tex}

\input{\figdir representative-results-control.tex}
\input{\figdir representative-results-overt.tex}

\subsubsection{Transformation parameters}

\input{\figdir matrix-loss.tex}

Ground truth parameters were compared to those predicted by the best performing network (\bestnetwork{}) via correlation, and by Bland Altman plots \hl{(Fig.~\ref{fig:matrix-loss})}.
It is notable that ground truth transformation parameters (particularly rotation and scale) were not uniformly distributed between views.
Nonrandom rotation is to be expected from the fact that the positioning of the patient in the scanner, the protocol for determining imaging planes, the placement of the heart in the chest, and the relationship between imaging planes are all themselves nonrandom; nonrandom scale is likewise to be expected from the variable size of the anatomical structures visible in each view.

Predicted horizontal translation, vertical translation, and rotation parameters were all highly correlated with ground truth ($R \approx~0.95$, $p < 0.0001$ for all), with the predicted parameters slightly under-estimating the ground truth (slope $\approx~0.87$ for all).  \hl{S}ystematic bias was \hl{not} evident on visual inspection of the Bland-Altman plots; $95\%$ of translation errors were within $\pm 0.07$ (in normalized image coordinates), and $95\%$ of rotation errors were within $\pm 0.63$ (in radians).
Of the $5\%$ of cases which were outside these bounds, the vast majority were long axis (\HLA{} or \VLA{}) views.
This is perhaps not surprising since each patient contributed three \SA{} views, but only two long axis views.

Compared with translation and rotation, correlation between ground truth and predicted scale was slightly lower, though still good ($R = 0.88$, $p < 0.0001$); predicted scale again slightly underestimated ground truth scale ($s = 0.71\hat{s} + 0.16$).
There \hl{wa}s a marked decrease in network performance above approximately $\hat{s} = 0.7$.
This may indicate the importance of context information to the network\hl{.
H}owever, it should be noted that the decrease in performance is accompanied by a sharp decrease in the frequency of cases, and so may also be the result of an insufficient number of samples in the dataset.

\subsubsection{Failure cases}

\input{\figdir failure.tex}

Occasional failure cases were observed, a selection of which are shown in Fig.~\ref{fig:failure}.
Each of these failure cases has one or more features which could logically explain the failure.
The leftmost column shows an apical \SA{} slice from a severely hypertrophied patient.
Patients with such severe disease were relatively uncommon in the dataset, perhaps causing the network to split its attention between the heart and a second ``candidate structure'' (the cardia of the stomach).
The center-left column shows a second apical \SA{} slice from a different subject, where the right ventricle was incorrectly segmented.
The signal intensity in this image was low relative to the other patients in the cohort, resulting in a very high contrast image after histogram equalization.
The center-right and rightmost columns show long axis views from a patient with a particularly high resolution scan, where the heart occupies the vast majority of the image, with very little context information.
In both cases, catastrophic segmentation error follows failure to properly reorient the image into a canonical orientation.
However, it should be emphasized that this post hoc reasoning is speculative; we cannot state a definitive causal relationship between these features and the resulting failures.

\hl{
\subsection{2017 MICCAI \miccaidata{} dataset}
\input{\tabdir acdc.tex}
\citet{Isensee2017} represents the state-of-the-art network in terms of segmentation accuracy on the \miccaidata{} leaderboard; this same group has since released an unpublished revision\footnote{https://arxiv.org/abs/1707.00587v2} with improved results \citep{Isensee2018}.
To match their methods, we retrained the Network B variant of \omeganet{} from scratch using five-fold cross-validation on the provided dataset~(each patient only appears in \emph{one} fold).
Single model segmentation accuracy is reported for \omeganet{}, \citet{Isensee2017}, and \citet{Isensee2018} in Table~\ref{tab:acdc}.
Compared with \citet{Isensee2017}, our results give higher \IoU{} for all foreground classes: \LV{} bloodpool ($\ACDCONJLVBP{}$ vs $\ACDCFIOJLVBP{}$), \RV{} bloodpool ($\ACDCONJRVBP{}$ vs $\ACDCFIOJRVBP{}$), and \LV{} myocardium ($\ACDCONJLVMY{}$ vs $\ACDCFIOJLVMY{}$).
Compared with \citet{Isensee2018}, our results give higher \IoU{} for \LV{} bloodpool ($\ACDCONJLVBP{}$ vs $\ACDCFINJLVBP{}$) and \RV{} bloodpool ($\ACDCONJRVBP{}$ vs $\ACDCFINJRVBP{}$), but lower \IoU{} for \LV{} myocardium ($\ACDCONJLVMY{}$ vs $\ACDCFINJLVMY{}$).
}

%% file: tables/table_2.tex
% {A,B,C,D}{S,H,V}{M,H,L}{Ze,On,Tw,Th}
% {Network}{View}{Quartile}{Hourglass}
\renewcommand{\captiontitle}{Network performance with reference to ground truth}
\begin{sidewaystable*}
\begin{center}
\begin{tabular}{ccccccc} \hline
\toprule
Name      & View   & \UNet{} 0                 & \UNet{} 1                 & \UNet{} 2                 & \UNet{} 3                 \\
\midrule
Network A & All    & $\AAMZe~[\AALZe,~\AAHZe]$ & --                        & --                        & --                        \\
          & \SA{}  & $\ASMZe~[\ASLZe,~\ASHZe]$ & --                        & --                        & --                        \\
          & \HLA{} & $\AHMZe~[\AHLZe,~\AHHZe]$ & --                        & --                        & --                        \\
          & \VLA{} & $\AVMZe~[\AVLZe,~\AVHZe]$ & --                        & --                        & --                        \\
\midrule
Network B & All    & $\BAMZe~[\BALZe,~\BAHZe]$ & $\BAMOn~[\BALOn,~\BAHOn]$ & --                        & --                        \\
          & \SA{}  & $\BSMZe~[\BSLZe,~\BSHZe]$ & $\BSMOn~[\BSLOn,~\BSHOn]$ & --                        & --                        \\
          & \HLA{} & $\BHMZe~[\BHLZe,~\BHHZe]$ & $\BHMOn~[\BHLOn,~\BHHOn]$ & --                        & --                        \\
          & \VLA{} & $\BVMZe~[\BVLZe,~\BVHZe]$ & $\BVMOn~[\BVLOn,~\BVHOn]$ & --                        & --                        \\
\midrule
Network C & All    & $\CAMZe~[\CALZe,~\CAHZe]$ & $\CAMOn~[\CALOn,~\CAHOn]$ & $\CAMTw~[\CALTw,~\CAHTw]$ & --                        \\
          & \SA{}  & $\CSMZe~[\CSLZe,~\CSHZe]$ & $\CSMOn~[\CSLOn,~\CSHOn]$ & $\CSMTw~[\CSLTw,~\CSHTw]$ & --                        \\
          & \HLA{} & $\CHMZe~[\CHLZe,~\CHHZe]$ & $\CHMOn~[\CHLOn,~\CHHOn]$ & $\CHMTw~[\CHLTw,~\CHHTw]$ & --                        \\
          & \VLA{} & $\CVMZe~[\CVLZe,~\CVHZe]$ & $\CVMOn~[\CVLOn,~\CVHOn]$ & $\CVMTw~[\CVLTw,~\CVHTw]$ & --                        \\
\midrule
Network D & All    & $\DAMZe~[\DALZe,~\DAHZe]$ & $\DAMOn~[\DALOn,~\DAHOn]$ & $\DAMTw~[\DALTw,~\DAHTw]$ & $\DAMTh~[\DALTh,~\DAHTh]$ \\
          & \SA{}  & $\DSMZe~[\DSLZe,~\DSHZe]$ & $\DSMOn~[\DSLOn,~\DSHOn]$ & $\DSMTw~[\DSLTw,~\DSHTw]$ & $\DSMTh~[\DSLTh,~\DSHTh]$ \\
          & \HLA{} & $\DHMZe~[\DHLZe,~\DHHZe]$ & $\DHMOn~[\DHLOn,~\DHHOn]$ & $\DHMTw~[\DHLTw,~\DHHTw]$ & $\DHMTh~[\DHLTh,~\DHHTh]$ \\
          & \VLA{} & $\DVMZe~[\DVLZe,~\DVHZe]$ & $\DVMOn~[\DVLOn,~\DVHOn]$ & $\DVMTw~[\DVLTw,~\DVHTw]$ & $\DVMTh~[\DVLTh,~\DVHTh]$ \\
\bottomrule
\end{tabular}
\caption[\captiontitle{}]{\captiontitle{}.
\hl{
Weighted foreground \IoU{} for each network variant is calculated for all views combined and for each view separately, and is reported as median [interquartile range].
}
Network performance was highest in the \SA{} view and lowest in the \HLA{} view, though these differences are small.
}
\label{tab:architectureaccuracy}
\end{center}
\end{sidewaystable*}

%% file: figures/hourglass-accuracy.tex
\renewcommand{\captiontitle}{Weighted foreground \IoU{} for architecture and depth}
\begin{figure}
\begin{center}
\includegraphics[width=0.5\textwidth]{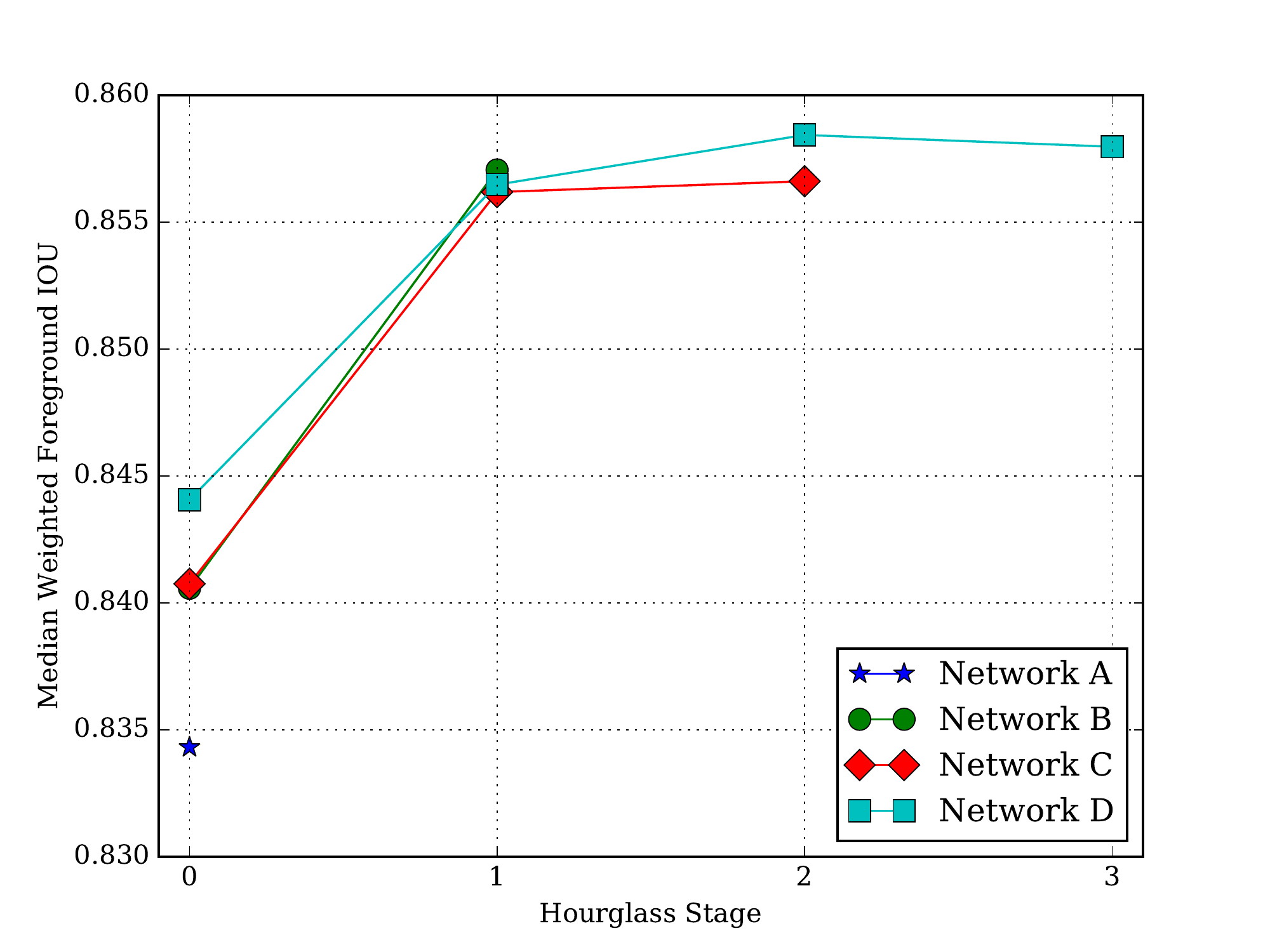}
\caption[\captiontitle]{\captiontitle{}.}
\label{fig:hourglass-accuracy}
\end{center}
\end{figure}

%% file: figures/histograms.tex
\renewcommand{\captiontitle}{Histograms of \IoU{} for each view and class}
\begin{figure*}
\begin{center}
\includegraphics[width=1.0\textwidth]{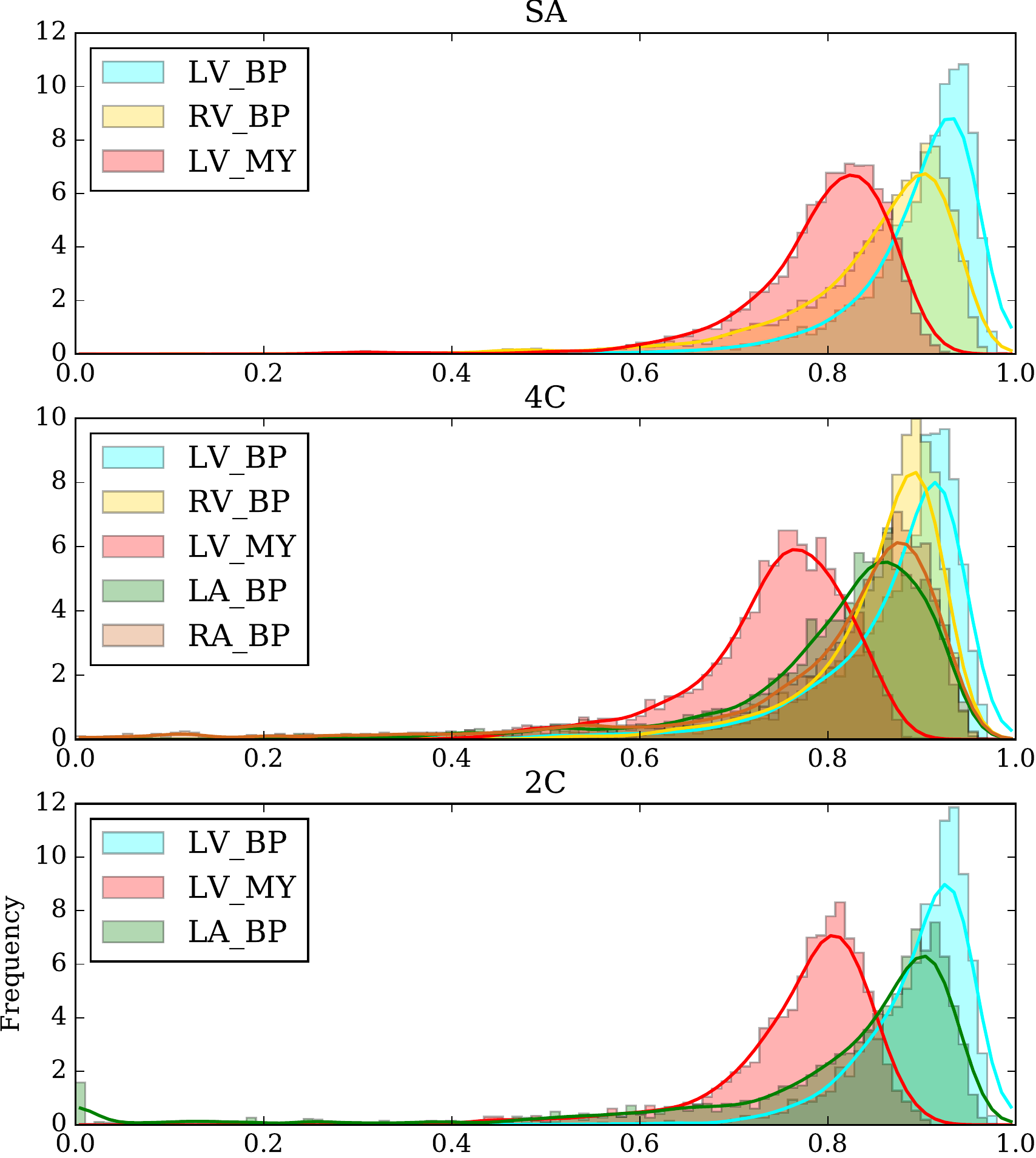}
\caption[\captiontitle]{\captiontitle{}.  Performance relative to manual segmentation was highest in the \SA{} view and lowest in the \HLA{} view, though the differences are small. \\
\hl{
LV\_BP: left ventricular blood pool; RV\_BP: right ventricular blood pool; LV\_MY: left ventricular myocardium; LA\_BP: left atrial blood pool; RA\_BP: right atrial blood pool.}
}
\label{fig:histograms}
\end{center}
\end{figure*}

%% file: figures/roc.tex
\renewcommand{\captiontitle}{Precision-Recall curve}
\begin{figure}
\begin{center}
\includegraphics[width=0.5\textwidth]{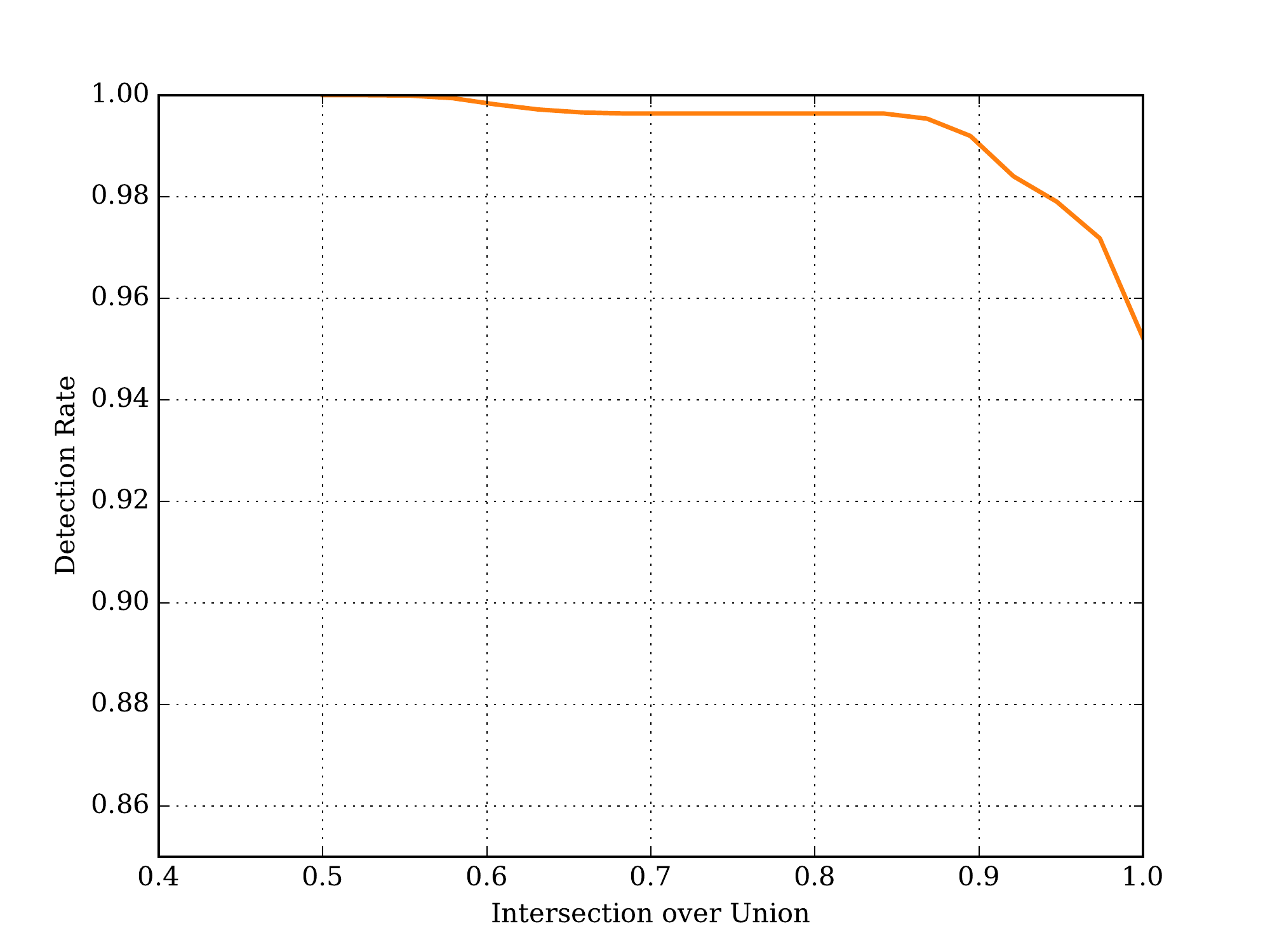}
\caption[\captiontitle]{\captiontitle{}.
}
\label{fig:roc}
\end{center}
\end{figure}

%% file: figures/representative-results-control.tex
\renewcommand{\captiontitle}{Representative segmentation results in healthy control subjects}
\begin{figure*}
\begin{center}

\setlength{\tabcolsep}{1pt}

\begin{tabular}{ccccc}

\toprule
\SA{} (Base) & \SA{} (Midslice) & \SA{} (Apex) & \HLA{} & \VLA{} \\
\midrule

\multicolumn{5}{c}{Original Untransformed Image: $\image$} \\

\includegraphics[width=0.19\textwidth]{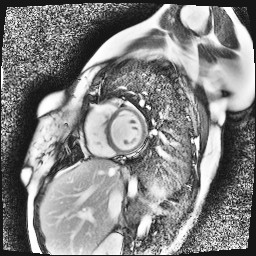} &
\includegraphics[width=0.19\textwidth]{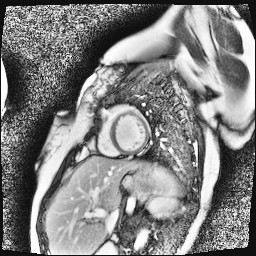} &
\includegraphics[width=0.19\textwidth]{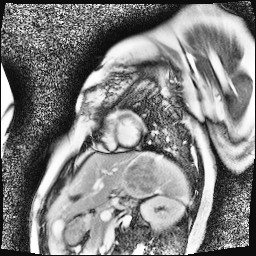} &
\includegraphics[width=0.19\textwidth]{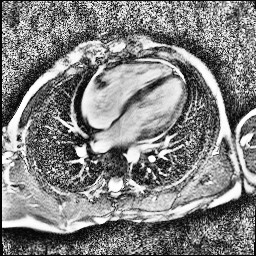} &
\includegraphics[width=0.19\textwidth]{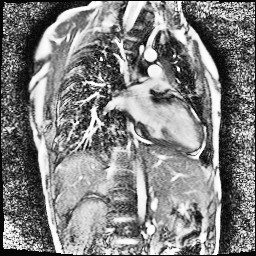} \\

\multicolumn{5}{c}{Structure \hl{Localization}} \\

\includegraphics[width=0.19\textwidth]{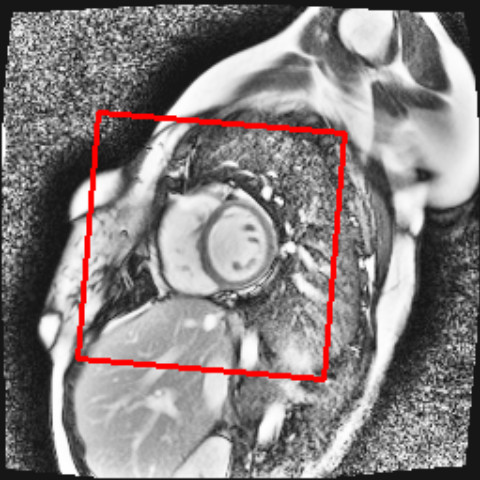} &
\includegraphics[width=0.19\textwidth]{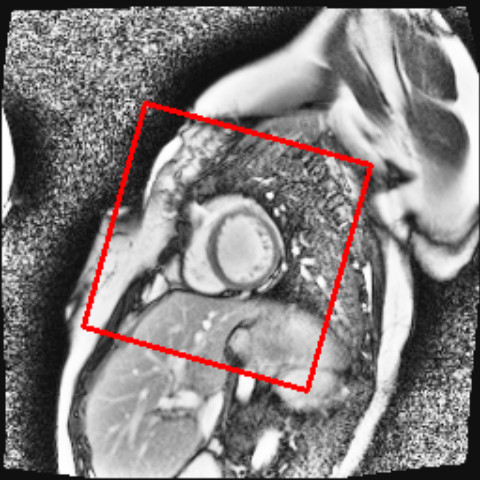} &
\includegraphics[width=0.19\textwidth]{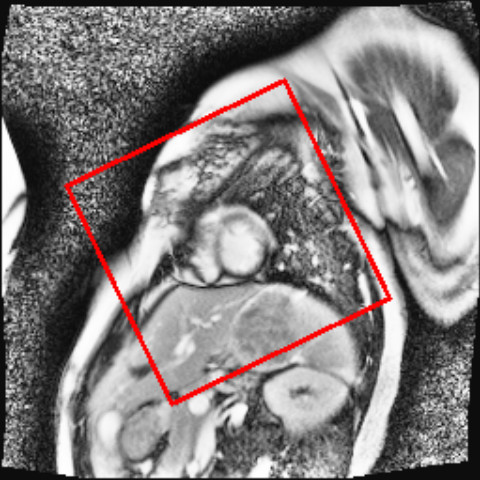} &
\includegraphics[width=0.19\textwidth]{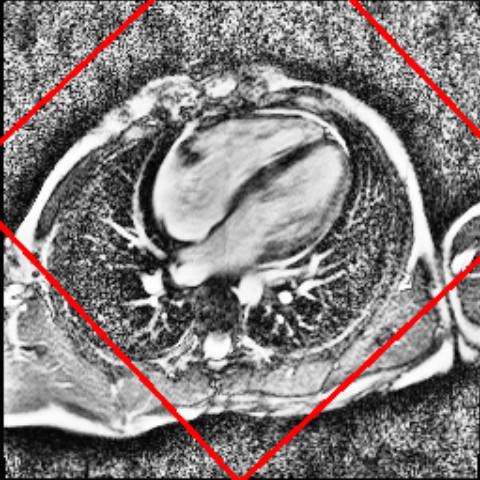} &
\includegraphics[width=0.19\textwidth]{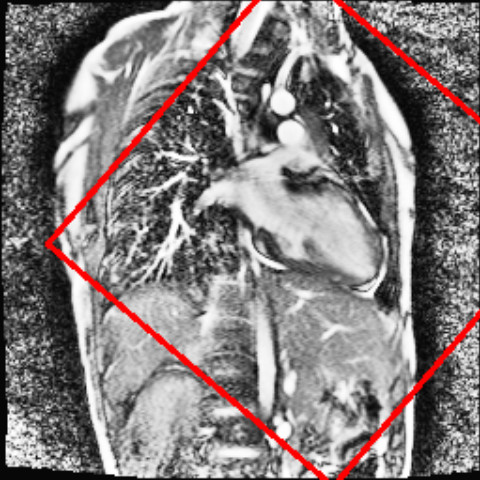} \\

\multicolumn{5}{c}{Predicted Segmentation in Canonical Orientation: $S^\prime$} \\

\includegraphics[width=0.19\textwidth]{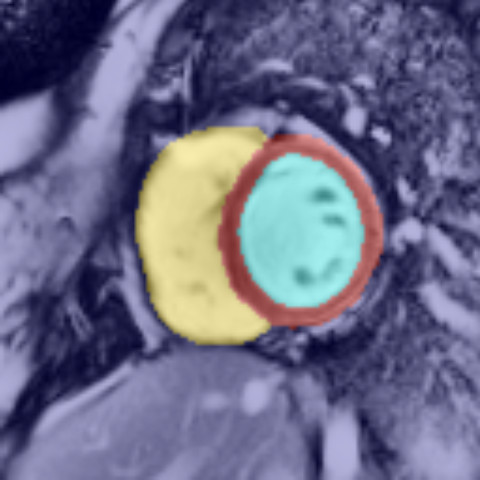} &
\includegraphics[width=0.19\textwidth]{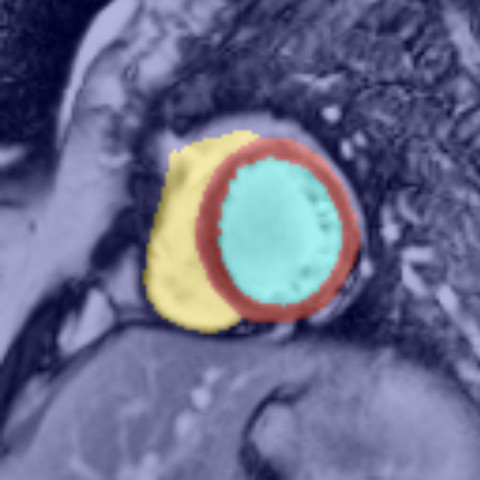} &
\includegraphics[width=0.19\textwidth]{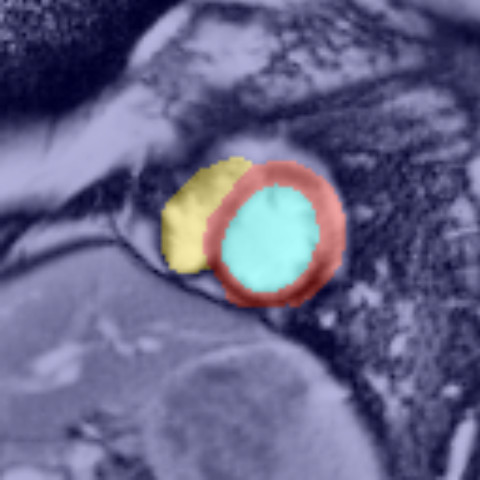} &
\includegraphics[width=0.19\textwidth]{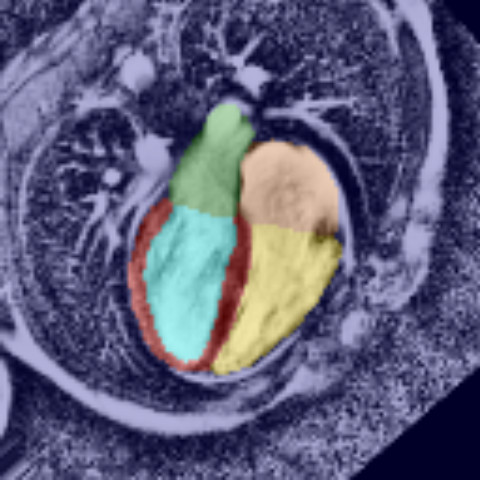} &
\includegraphics[width=0.19\textwidth]{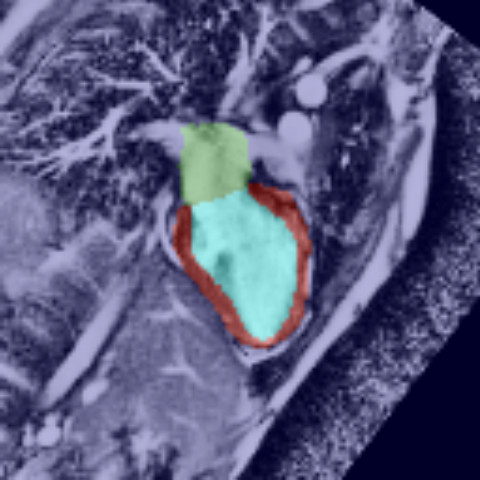} \\
\bottomrule

\multicolumn{5}{c}{Ground Truth Segmentation in Canonical Orientation: $\hat{S}^\prime$} \\

\includegraphics[width=0.19\textwidth]{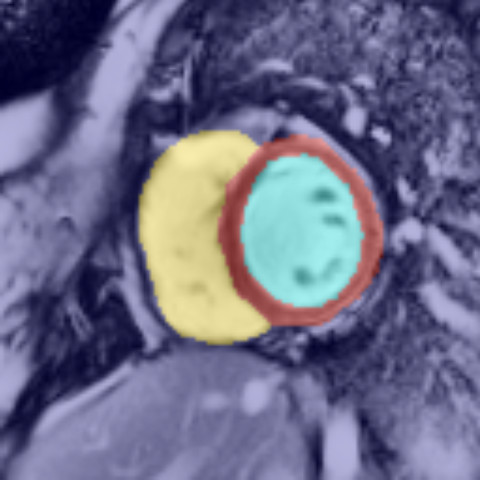} &
\includegraphics[width=0.19\textwidth]{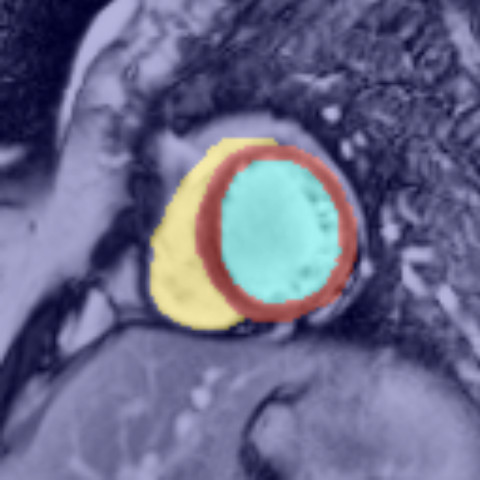} &
\includegraphics[width=0.19\textwidth]{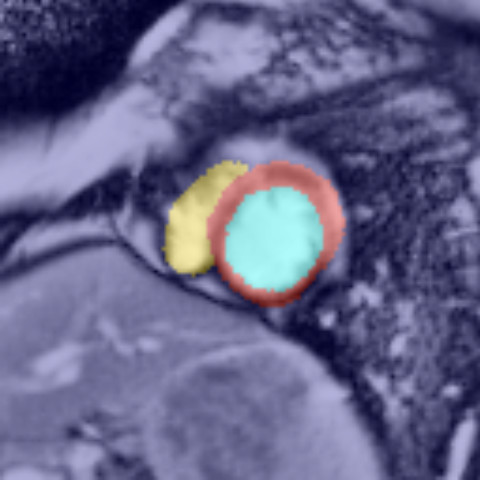} &
\includegraphics[width=0.19\textwidth]{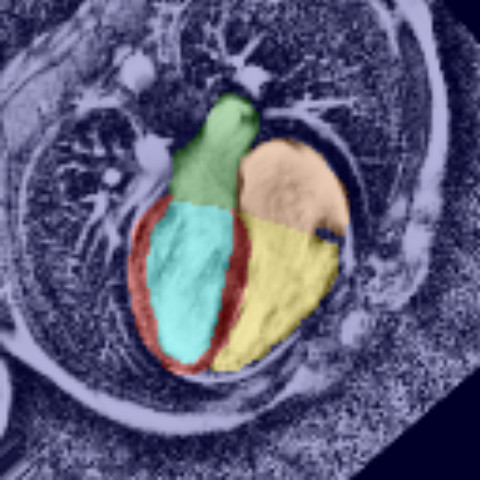} &
\includegraphics[width=0.19\textwidth]{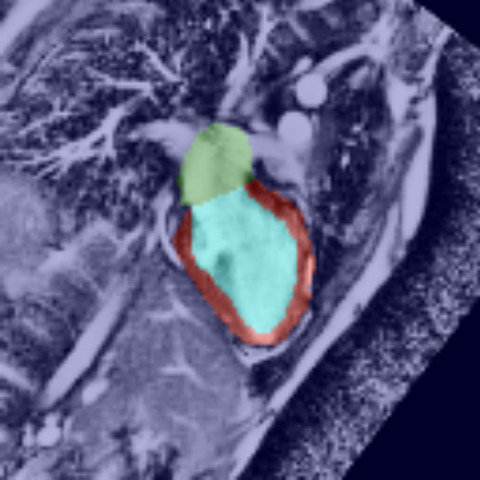} \\
\bottomrule

\end{tabular}

\caption[\captiontitle]{\captiontitle{}.  See text for discussion.}
\label{fig:representative-results-control}
\end{center}
\end{figure*}

%% file: figures/representative-results-overt.tex
\renewcommand{\captiontitle}{Representative segmentation results in patients with overt \HCM{}}
\begin{figure*}
\begin{center}

\setlength{\tabcolsep}{1pt}

\begin{tabular}{ccccc}

\toprule
\SA{} (Base) & \SA{} (Midslice) & \SA{} (Apex) & \HLA{} & \VLA{} \\
\midrule

\multicolumn{5}{c}{Original Untransformed Image: $\image$} \\

\includegraphics[width=0.19\textwidth]{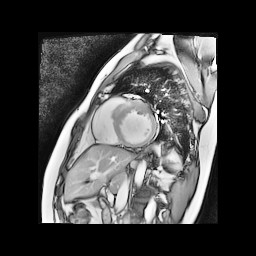} &
\includegraphics[width=0.19\textwidth]{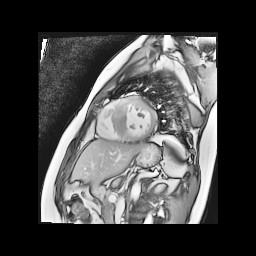} &
\includegraphics[width=0.19\textwidth]{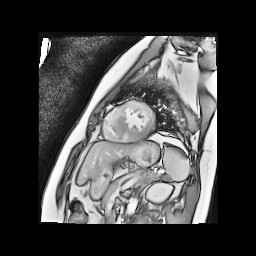} &
\includegraphics[width=0.19\textwidth]{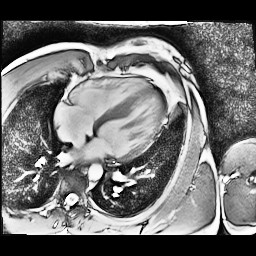} &
\includegraphics[width=0.19\textwidth]{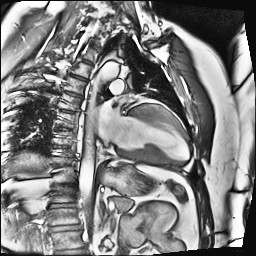} \\

\multicolumn{5}{c}{Structure \hl{Localization}} \\

\includegraphics[width=0.19\textwidth]{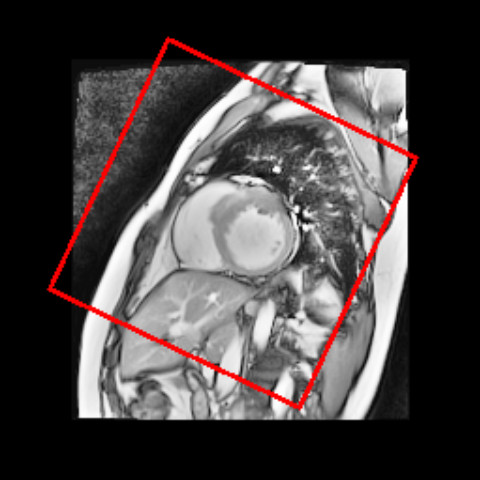} &
\includegraphics[width=0.19\textwidth]{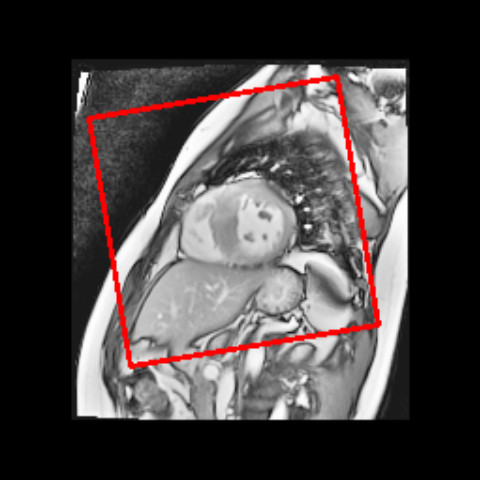} &
\includegraphics[width=0.19\textwidth]{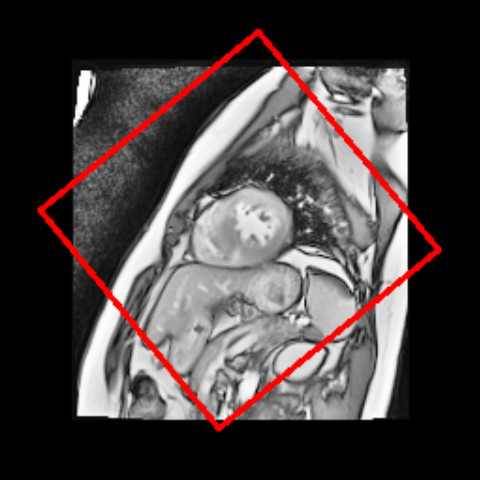} &
\includegraphics[width=0.19\textwidth]{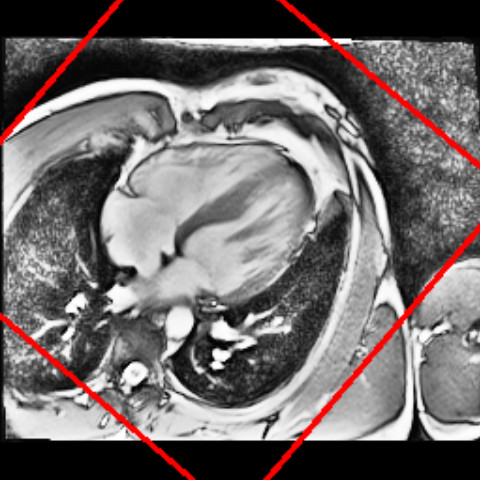} &
\includegraphics[width=0.19\textwidth]{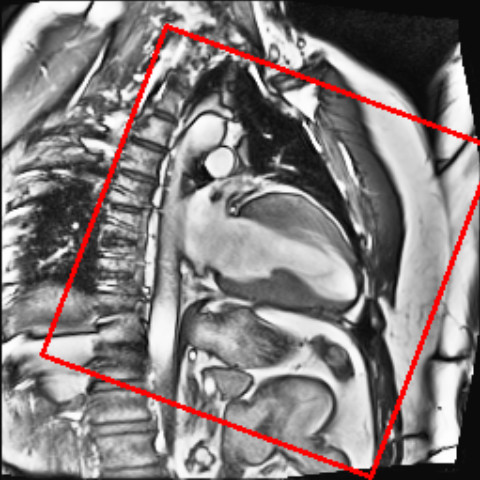} \\

\multicolumn{5}{c}{Predicted Segmentation in Canonical Orientation: $S^\prime$} \\

\includegraphics[width=0.19\textwidth]{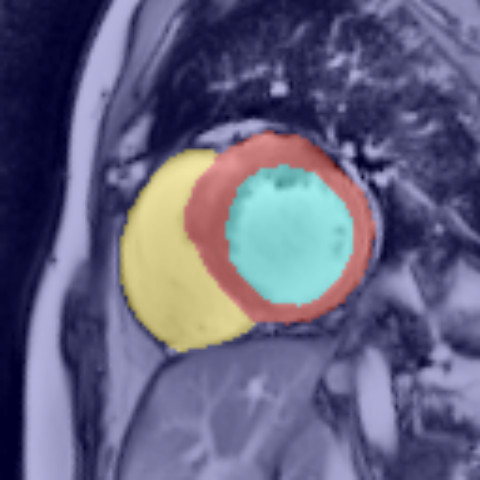} &
\includegraphics[width=0.19\textwidth]{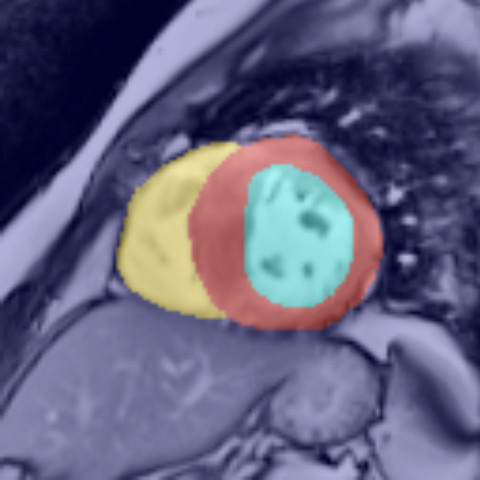} &
\includegraphics[width=0.19\textwidth]{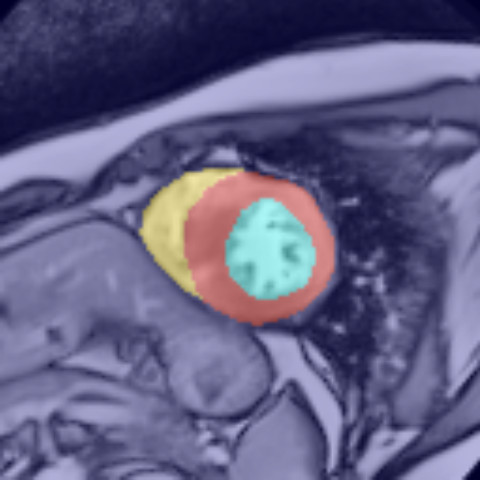} &
\includegraphics[width=0.19\textwidth]{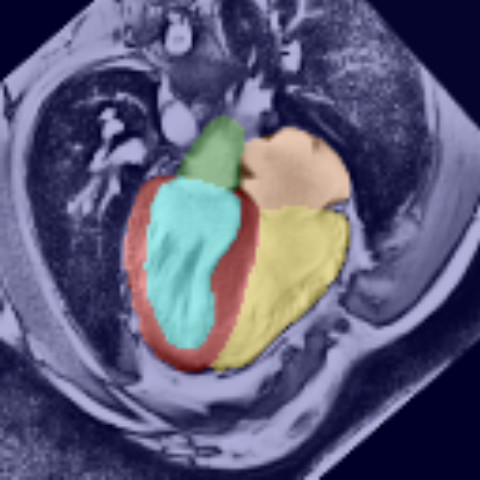} &
\includegraphics[width=0.19\textwidth]{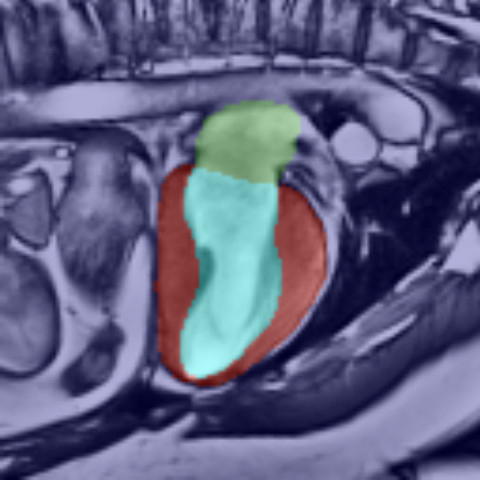} \\
\bottomrule

\multicolumn{5}{c}{Ground Truth Segmentation in Canonical Orientation: $\hat{S}^\prime$} \\

\includegraphics[width=0.19\textwidth]{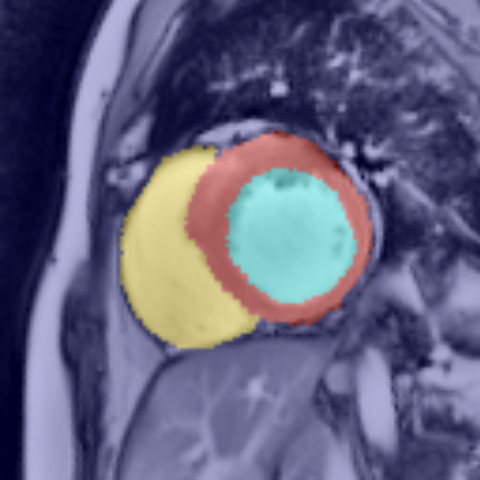} &
\includegraphics[width=0.19\textwidth]{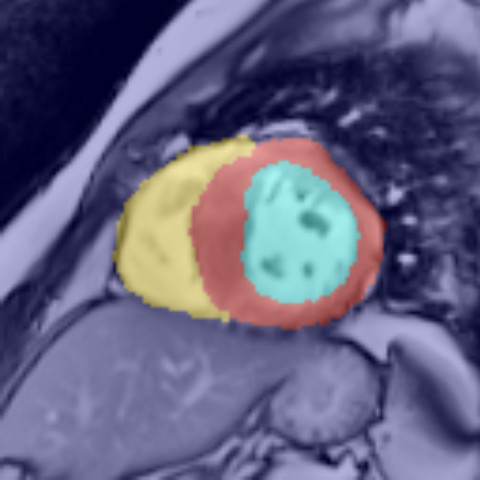} &
\includegraphics[width=0.19\textwidth]{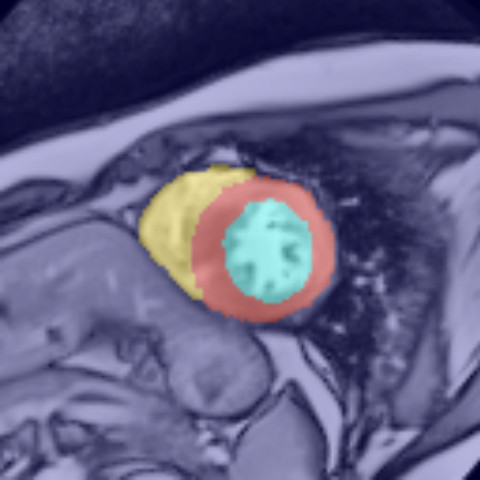} &
\includegraphics[width=0.19\textwidth]{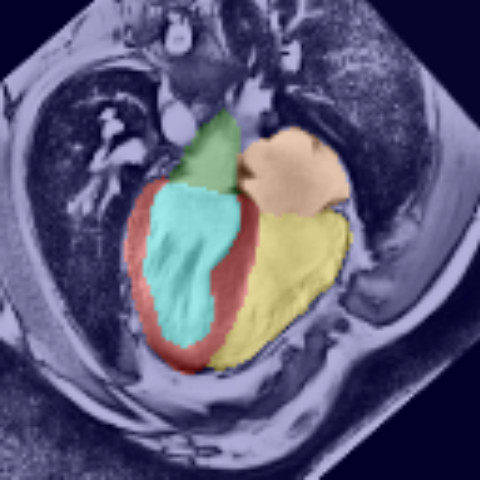} &
\includegraphics[width=0.19\textwidth]{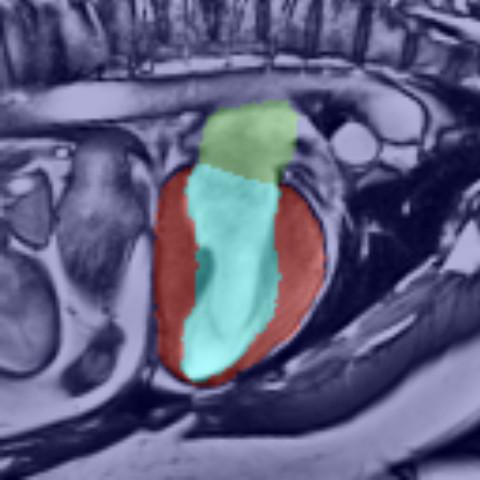} \\
\bottomrule

\end{tabular}

\caption[\captiontitle]{\captiontitle{}.  See text for discussion.}
\label{fig:representative-results-overt}
\end{center}
\end{figure*}

%% file: figures/matrix-loss.tex
\renewcommand{\captiontitle}{Transformation errors}
\begin{sidewaysfigure*}
\begin{center}
\includegraphics[width=1.0\textwidth]{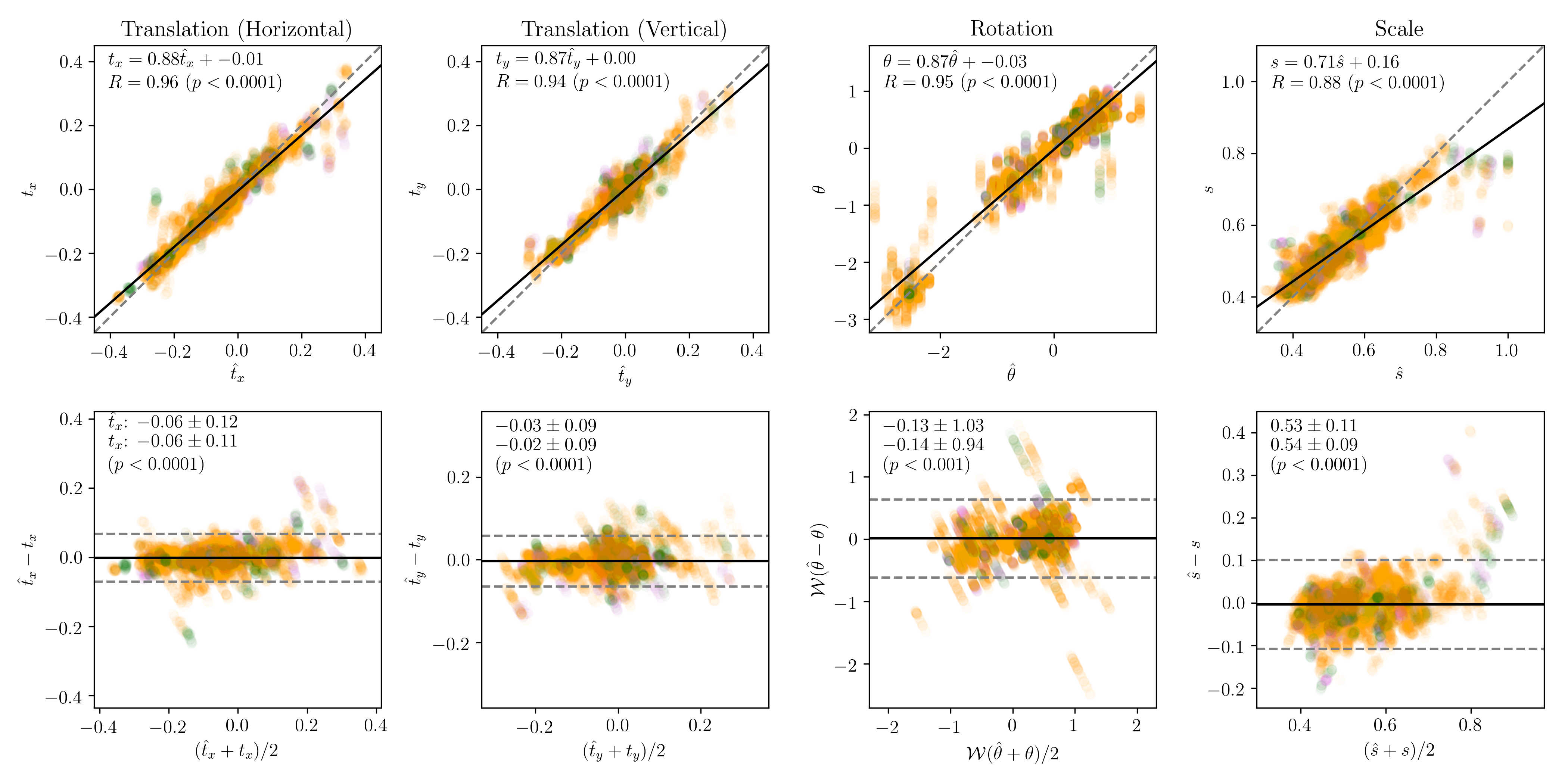}
\caption[\captiontitle]{\captiontitle{}.
Correlation (top) and Bland-Altman (bottom) plots comparing predicted and ground truth transformation parameters.
\SA{}, \HLA{}, and \VLA{} errors are represented by orange, green, and purple points, respectively (points have been rendered partially translucent to aid interpretation of densely occupied regions).
In the correlation plots, the best-fit trendline is represented as a solid black line, and the ideal trendline ($y = x$) is represented as a dashed, gray line.
The equation of the trendline and the Pearson correlation coefficient $R$ are also given.
In the Bland-Altman plots, the mean difference is represented as a solid black horizontal line, and the limits $\pm 1.96$ standard deviations are represented as a dashed gray horizontal line.
}
\label{fig:matrix-loss}
\end{center}
\end{sidewaysfigure*}

%% file: figures/failure.tex
\renewcommand{\captiontitle}{Selected failure cases from \CNN{} segmentation}
\begin{figure*}
\begin{center}

\setlength{\tabcolsep}{1pt}

\begin{tabular}{cccc}

\toprule
\SA{} (Apex) & \SA{} (Apex) & \HLA{} & \VLA{} \\
\midrule

\multicolumn{4}{c}{Original Untransformed Image: $\image$} \\

\includegraphics[width=0.19\textwidth]{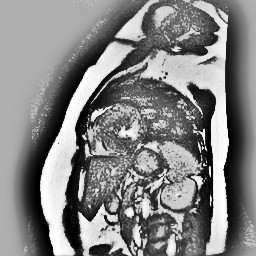} &
\includegraphics[width=0.19\textwidth]{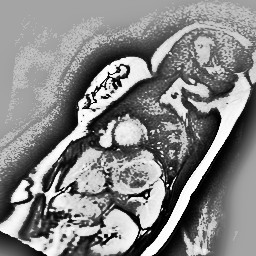} &
\includegraphics[width=0.19\textwidth]{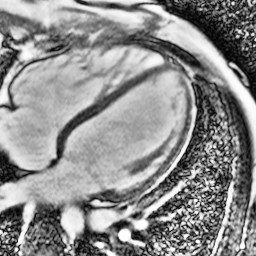} &
\includegraphics[width=0.19\textwidth]{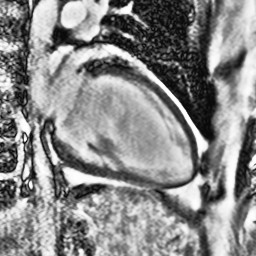} \\

\multicolumn{4}{c}{Structure \hl{Localization}} \\

\includegraphics[width=0.19\textwidth]{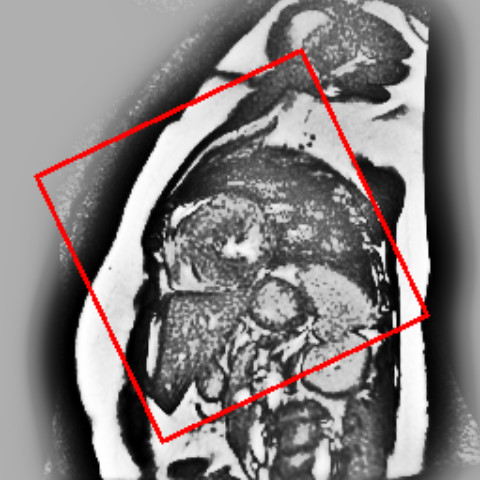} &
\includegraphics[width=0.19\textwidth]{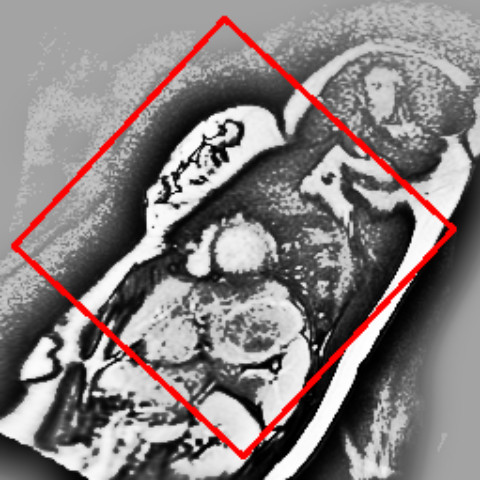} &
\includegraphics[width=0.19\textwidth]{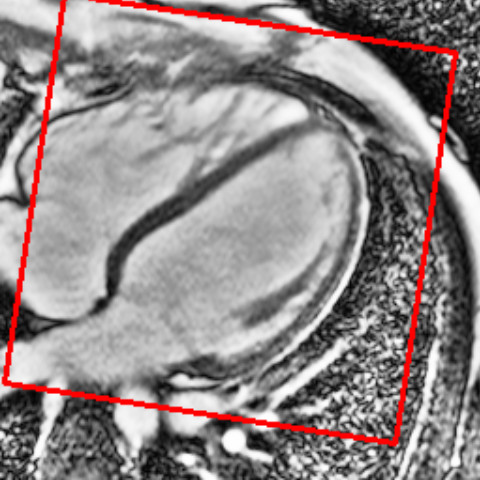} &
\includegraphics[width=0.19\textwidth]{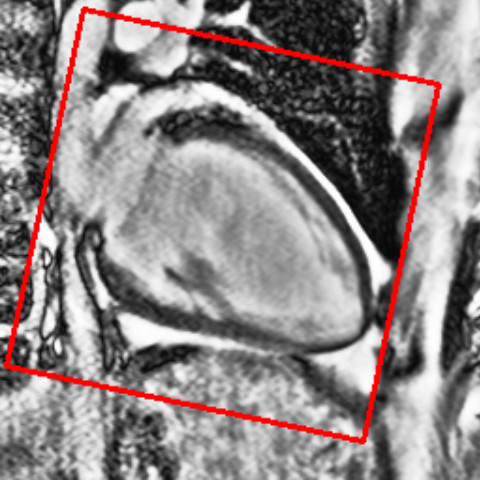} \\

\multicolumn{4}{c}{Predicted Segmentation in Canonical Orientation: $S^\prime$} \\

\includegraphics[width=0.19\textwidth]{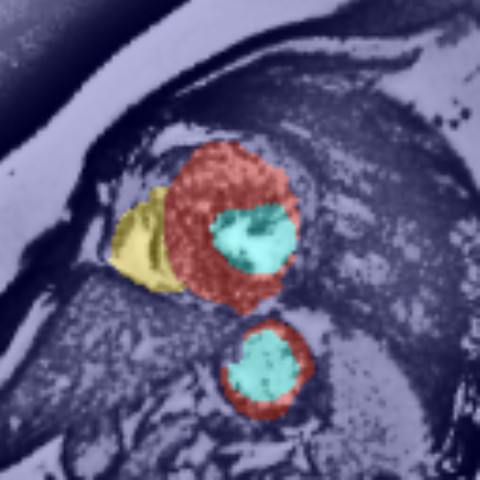} &
\includegraphics[width=0.19\textwidth]{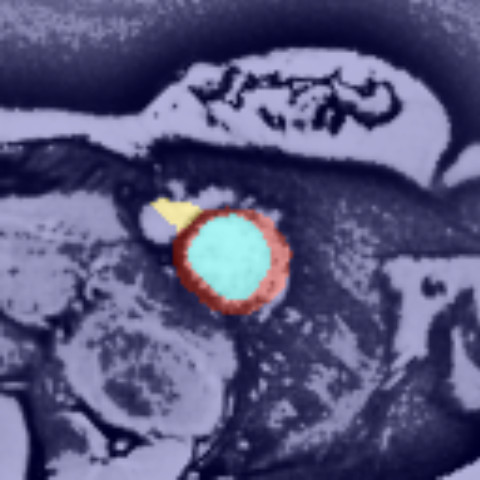} &
\includegraphics[width=0.19\textwidth]{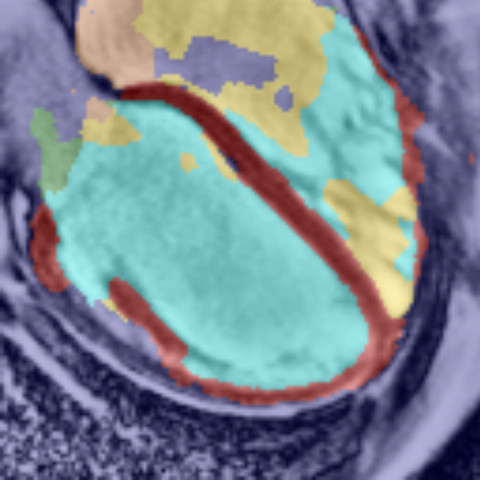} &
\includegraphics[width=0.19\textwidth]{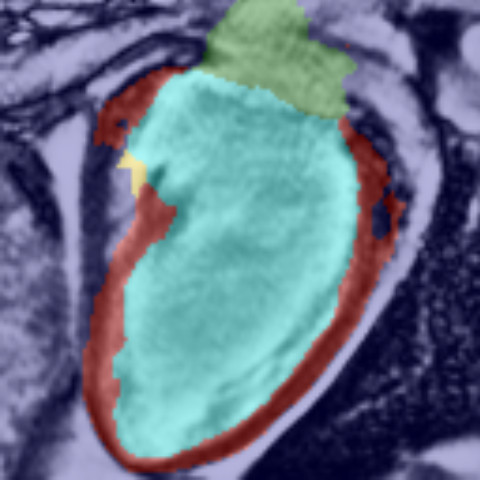} \\
\bottomrule

\multicolumn{4}{c}{Ground Truth Segmentation in Canonical Orientation: $\hat{S}^\prime$} \\

\includegraphics[width=0.19\textwidth]{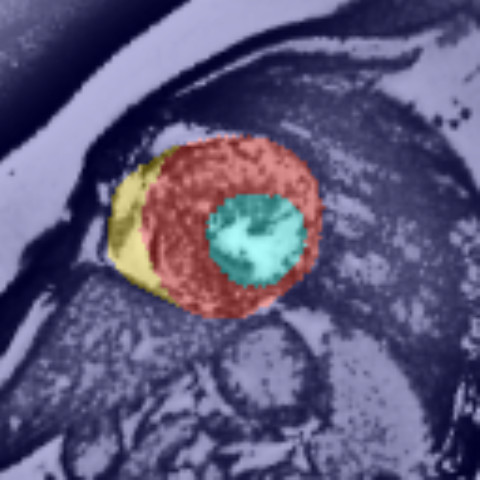} &
\includegraphics[width=0.19\textwidth]{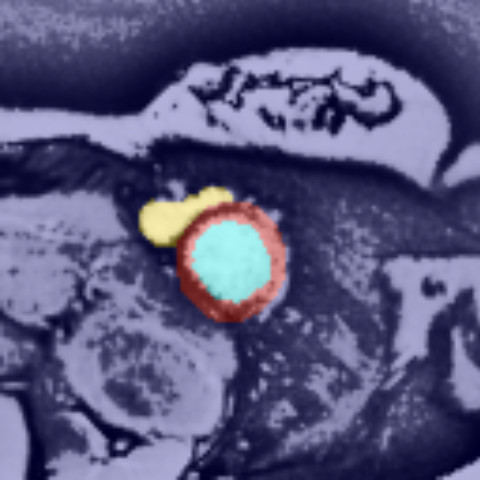} &
\includegraphics[width=0.19\textwidth]{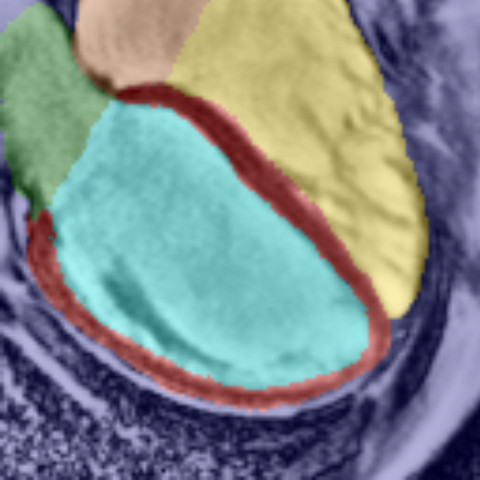} &
\includegraphics[width=0.19\textwidth]{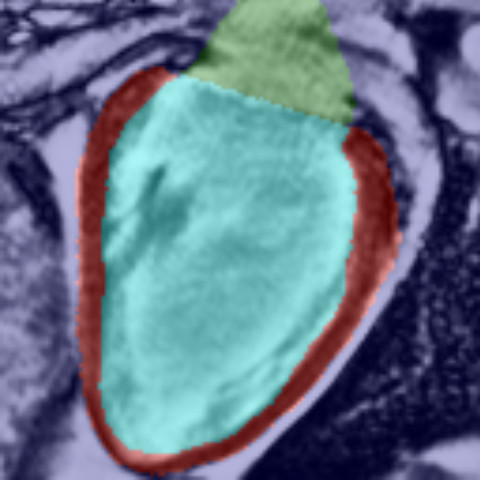} \\
\bottomrule

\end{tabular}

\caption[\captiontitle]{\captiontitle{}.  See text for discussion.}
\label{fig:failure}
\end{center}
\end{figure*}

%% file: tables/acdc.tex
\renewcommand{\captiontitle}{Segmentation accuracy on the 2017 MICCIA \miccaidata{} dataset}
\begin{table*}
\hl{
\begin{center}
\begin{tabular}{lccc} \hline
\toprule
Structure           & \LV{} Bloodpool & \RV{} Bloodpool & \LV{} Myocardium \\
\midrule
\multicolumn{4}{c}{Jaccard Index (\IoU{})} \\
\midrule
\omeganet{}         & \textbf{\ACDCONJLVBP{}}  & \textbf{\ACDCONJRVBP{}}  & \ACDCONJLVMY{}           \\
\citet{Isensee2018} & \ACDCFINJLVBP{}          & \ACDCFINJRVBP{}          & \textbf{\ACDCFINJLVMY{}} \\
\citet{Isensee2017} & \ACDCFIOJLVBP{}          & \ACDCFIOJRVBP{}          & \ACDCFIOJLVMY{}          \\
\midrule
\multicolumn{4}{c}{Dice Coefficient} \\
\midrule
\omeganet{}         & \textbf{\ACDCONDLVBP{}}  & \textbf{\ACDCONDRVBP{}}  & \ACDCONDLVMY{}           \\
\citet{Isensee2018} & \ACDCFINDLVBP{}          & \ACDCFINDRVBP{}          & \textbf{\ACDCFINDLVMY{}} \\
\citet{Isensee2017} & \ACDCFIODLVBP{}          & \ACDCFIODRVBP{}          & \ACDCFIODLVMY{}          \\
\bottomrule
\end{tabular}
\caption[\captiontitle{}]{
\hl{
\captiontitle{}.
Segmentation accuracy is reported as Dice coefficient in the \miccaidata{} challenge, but as \IoU{} elsewhere in this work; therefore, both are reported here.
(Note that $\mathrm{Dice} = 2 * \mathrm{\IoU{}} / (1 + \mathrm{\IoU{}})$).
Results are reported for the Network B variant of the \omeganet{}; for the results by \citet{Isensee2017} published in STACOM; and for the same group's unpublished \url{arxiv.org} revision \citet{Isensee2018}.
Boldface formatting indicates the best performing model for each foreground class.
}
}
\label{tab:acdc}
\end{center}
}
\end{table*}

%% file: text/discussion.tex
In this work, we have presented the \omeganet{}: a novel deep convolutional neural network (\CNN{}) architecture for \hl{localization}, orientation alignment, and segmentation.
We have applied this network to the task of fully automatic whole-heart segmentation and simultaneous transformation into the ``canonical'' clinical view, 
which has the potential to greatly simplify downstream analyses of \SSFP{} \CMR{} images.
The network was trained end-to-end from scratch to segment five foreground classes (the four cardiac chambers plus the \LV{} myocardium) in three views (\SA{}, \HLA{}, 
and \VLA{}), without providing prior knowledge of the view being segmented.
The dataset was highly heterogeneous from the standpoint of anatomical variation, including both healthy subjects and patients with overt hypertrophic cardiomyopathy.
Data was acquired from both 1.5-T and 3-T magnets as part of a multicenter trial involving 10 institutions.
In cross-validation experiments, the network performed well in predicting both the parameters of the transformation, and the cardiac segmentation.

\hl{
\omeganet{} also achieved state-of-the-art performance on the publicly available 2017 MICCAI \miccaidata{} dataset in two of three classes.
Compared with our internal HCMNet dataset, \miccaidata{} contains a broader range of \LV{} and \RV{} pathologies, but only one clinical view, and fewer foreground classes.
Moreover, HCMNet was a multicenter study, whereas \miccaidata{} was acquired at a single center.
It is encouraging that \omeganet{} performed well on both datasets.

The prior state-of-the-art \citep{Isensee2017,Isensee2018} was achieved using an ensemble of 2D and 3D \UNet{}-inspired architectures, optimized for \emph{stacked} cine series.
Their method is therefore not generally applicable to \HLA{} and \VLA{} views, which are typically acquired as single slices.
Therefore, \omeganet{} outperformed \citet{Isensee2018} while remaining more general, and while providing \hl{localization} and orientation information not predicted by \citep{Isensee2017}.
}

The work is novel in \hl{four} principal ways.
First, this network predicts five foreground classes in three clinical views, which is a substantially more difficult problem than has been addressed previously in the literature \citep{Vigneault2017}.
Second, a spatial transformer network module \citep{Jaderberg2015} was used to rotate each view into a canonical orientation.
\hl{
\CNN{}s are neither rotation invariant nor equivariant, nor scale invariant.
From a technical standpoint, in theory this shortcoming can be addressed by acquiring very large datasets which adequately represent all possible rotations.
However, biomedical imaging datasets are expensive and time consuming both to acquire and to annotate, directly motivating this design decision.
By standardizing the orientation of the input to the \hl{final} segmentation module, we simplify the task of both the downstream network and the physician interpreting the images.}
Third, the proposed architecture takes loose inspiration from the cascaded classifier models proposed by \citet{Viola2001}, in that \UNet{} $0$ performs \hl{initial} segmentation (in order to predict transformation parameters), and the transformed image is then provided as input to a \hl{final} segmentation module (\UNet{}s $1$, $2$, and $3$).
\hl{
Last, by its design, \omeganet{} provides human-interpretable, intermediate outputs (an \hl{initial} segmentation and transformation parameters) in addition to the \hl{final} segmentation.
In doing so, we substantially increase the complexity and information predicted by the network compared to the \UNet{} architecture, but without adding concerns that \CNN{}s are ``black boxes'' whose internals cannot be adequately interrogated.
}

Although the dataset included three orthogonal cardiac planes and both healthy \hl{subjects and those with \LV{} pathology}, there remain potential opportunities to extend the dataset to more general scenarios.
First, other cardiac planes used in clinical practice (such as the axial, three-chamber, and \RV{} long axis views) should be added in future work.
It would also be useful and interesting to test this on other \CMR{} pulse sequences (such as gradient echo) and on additional modalities (i.e., cardiac computed tomography and echocardiography).
Moreover, it could also be interesting to apply this technique to other areas within biomedical image segmentation where \hl{localization}, reorientation, and segmentation are useful, such as in fetal imaging.
\hl{
Finally, we expect \omeganet{} to be useful in applications requiring the segmentation of multiple clinical planes, such as \CMR{} motion correction and slice alignment \citep{Sinclair2017}.
}

A variety of opportunities present themselves in terms of optimizing the \omeganet{} architecture.
For example, the network was trained to segment individual image frames, without spatial or temporal context; modifying the architecture to allow information sharing between temporal frames and spatial slices has the potential to increase accuracy and consistency.
The E-Net (``Efficient Net'') provides modifications to the \UNet{} blocks which increase computational and memory efficiency, while preserving accuracy \citep{Paszke2016}; these lessons have been applied successfully to cardiac segmentation \citep{Lieman-Sifry2017}, and could theoretically be applied here as well.

%% file: text/summary.tex
\hl{W}e have presented \omeganet{} (Omega-Net): a novel \CNN{} architecture for simultaneous \hl{localization}, transformation into a canonical orientation, and semantic segmentation.
First, an \hl{initial} segmentation is performed on the input image; second, the features learned during this \hl{initial} segmentation are used to predict the parameters needed to transform the input image into a canonical orientation; and third, a \hl{final} segmentation is performed on the transformed image.
The network was trained end-to-end from scratch \hl{on two different datasets.
On the HCMNet dataset,} \omeganet{} was trained to predict five foreground classes in three clinical views, constituting a substantially more challenging problem compared with prior work.
The trained network performed well in a cohort of both healthy subjects and patients with severe \LV{} pathology.
\hl{
A variant of the \omeganet{} network was trained from scratch on a publicly-available dataset, and achieved state-of-the-art performance in two of three segmenttion classes.
}
We believe this architecture represents a substantive advancement over prior approaches, with implications for biomedical image segmentation more generally.

%% file: text/acknowledgements.tex
D.M. Vigneault is supported by the NIH-Oxford Scholars Program and the NIH Intramural Research Program.
W. Xie is supported by the Google DeepMind Scholarship, and the EPSRC Programme Grant Seebibyte EP/M013774/1.

%% file: manuscript-omega-net.bbl
\begin{thebibliography}{32}
\expandafter\ifx\csname natexlab\endcsname\relax\def\natexlab#1{#1}\fi
\providecommand{\url}[1]{\texttt{#1}}
\providecommand{\href}[2]{#2}
\providecommand{\path}[1]{#1}
\providecommand{\DOIprefix}{doi:}
\providecommand{\ArXivprefix}{arXiv:}
\providecommand{\URLprefix}{URL: }
\providecommand{\Pubmedprefix}{pmid:}
\providecommand{\doi}[1]{\href{http://dx.doi.org/#1}{\path{#1}}}
\providecommand{\Pubmed}[1]{\href{pmid:#1}{\path{#1}}}
\providecommand{\bibinfo}[2]{#2}
\ifx\xfnm\undefined \def\xfnm[#1]{\unskip,\space#1}\fi
%Type = Unpublished
\bibitem[{Abadi et~al.(2016)Abadi, Barham, Chen, Chen, Davis, Dean, Devin,
  Ghemawat, Irving, Isard, Kudlur, Levenberg, Monga, Moore, Murray, Steiner,
  Tucker, Vasudevan, Warden, Wicke, Yu and Zheng}]{Abadi2016}
\bibinfo{author}{Abadi\xfnm[ M.]}, \bibinfo{author}{Barham\xfnm[ P.]},
  \bibinfo{author}{Chen\xfnm[ J.]}, \bibinfo{author}{Chen\xfnm[ Z.]},
  \bibinfo{author}{Davis\xfnm[ A.]}, \bibinfo{author}{Dean\xfnm[ J.]},
  \bibinfo{author}{Devin\xfnm[ M.]}, \bibinfo{author}{Ghemawat\xfnm[ S.]},
  \bibinfo{author}{Irving\xfnm[ G.]}, \bibinfo{author}{Isard\xfnm[ M.]},
  \bibinfo{author}{Kudlur\xfnm[ M.]}, \bibinfo{author}{Levenberg\xfnm[ J.]},
  \bibinfo{author}{Monga\xfnm[ R.]}, \bibinfo{author}{Moore\xfnm[ S.]},
  \bibinfo{author}{Murray\xfnm[ D.G.]}, \bibinfo{author}{Steiner\xfnm[ B.]},
  \bibinfo{author}{Tucker\xfnm[ P.]}, \bibinfo{author}{Vasudevan\xfnm[ V.]},
  \bibinfo{author}{Warden\xfnm[ P.]}, \bibinfo{author}{Wicke\xfnm[ M.]},
  \bibinfo{author}{Yu\xfnm[ Y.]}, \bibinfo{author}{Zheng\xfnm[ X.]}.
\newblock \bibinfo{title}{{TensorFlow: A system for large-scale machine
  learning}}; \bibinfo{year}{2016}.
\newblock \URLprefix \url{https://arxiv.org/pdf/1605.08695.pdf}.
  \href{http://arxiv.org/abs/arXiv:1605.08695v2}{\tt arXiv:arXiv:1605.08695v2}.
%Type = Misc
\bibitem[{Chollet(2015)}]{Chollet2015}
\bibinfo{author}{Chollet\xfnm[ F.]}.
\newblock \bibinfo{title}{{Keras}}.
\newblock \bibinfo{year}{2015}.
\newblock \URLprefix \url{https://github.com/fchollet/keras}.
%Type = Article
\bibitem[{Dieleman et~al.(2015)Dieleman, Willett and Dambre}]{Dieleman2015}
\bibinfo{author}{Dieleman\xfnm[ S.]}, \bibinfo{author}{Willett\xfnm[ K.W.]},
  \bibinfo{author}{Dambre\xfnm[ J.]}.
\newblock \bibinfo{title}{{Rotation-invariant convolutional neural networks for
  galaxy morphology prediction}}.
\newblock \bibinfo{journal}{Monthly Notices of the Royal Astronomical Society}
  \bibinfo{year}{2015};\bibinfo{volume}{450}(\bibinfo{number}{2}):\bibinfo{pages}{1441--1459}.
\newblock \DOIprefix\doi{10.1093/mnras/stv632}.
  \href{http://arxiv.org/abs/1503.07077}{\tt arXiv:1503.07077}.
%Type = Inproceedings
\bibitem[{He et~al.(2016)He, Zhang, Ren and Sun}]{He2016}
\bibinfo{author}{He\xfnm[ K.]}, \bibinfo{author}{Zhang\xfnm[ X.]},
  \bibinfo{author}{Ren\xfnm[ S.]}, \bibinfo{author}{Sun\xfnm[ J.]}.
\newblock \bibinfo{title}{{Deep Residual Learning for Image Recognition}}.
\newblock In: \bibinfo{booktitle}{CVPR}. \bibinfo{year}{2016}.
  \href{http://arxiv.org/abs/arXiv:1512.03385v1}{\tt arXiv:arXiv:1512.03385v1}.
%Type = Article
\bibitem[{Ho et~al.(2017)Ho, Day, Colan, Russell, Towbin, Sherrid, Canter,
  Jefferies, Murphy, Cirino, Abraham, Taylor, Mestroni, Bluemke, Jarolim, Shi,
  Sleeper, Seidman and Orav}]{Ho2017}
\bibinfo{author}{Ho\xfnm[ C.Y.]}, \bibinfo{author}{Day\xfnm[ S.M.]},
  \bibinfo{author}{Colan\xfnm[ S.D.]}, \bibinfo{author}{Russell\xfnm[ M.W.]},
  \bibinfo{author}{Towbin\xfnm[ J.A.]}, \bibinfo{author}{Sherrid\xfnm[ M.V.]},
  \bibinfo{author}{Canter\xfnm[ C.E.]}, \bibinfo{author}{Jefferies\xfnm[
  J.L.]}, \bibinfo{author}{Murphy\xfnm[ A.M.]}, \bibinfo{author}{Cirino\xfnm[
  A.L.]}, \bibinfo{author}{Abraham\xfnm[ T.P.]}, \bibinfo{author}{Taylor\xfnm[
  M.]}, \bibinfo{author}{Mestroni\xfnm[ L.]}, \bibinfo{author}{Bluemke\xfnm[
  D.A.]}, \bibinfo{author}{Jarolim\xfnm[ P.]}, \bibinfo{author}{Shi\xfnm[ L.]},
  \bibinfo{author}{Sleeper\xfnm[ L.A.]}, \bibinfo{author}{Seidman\xfnm[ C.E.]},
  \bibinfo{author}{Orav\xfnm[ E.J.]}.
\newblock \bibinfo{title}{{The Burden of Early Phenotypes and the Influence of
  Wall Thickness in Hypertrophic Cardiomyopathy Mutation Carriers}}.
\newblock \bibinfo{journal}{JAMA Cardiology}
  \bibinfo{year}{2017};\bibinfo{volume}{2}(\bibinfo{number}{4}):\bibinfo{pages}{419--428}.
\newblock \URLprefix
  \url{http://cardiology.jamanetwork.com/article.aspx?doi=10.1001/jamacardio.2016.5670
  http://cardiology.jamanetwork.com/article.aspx?doi=10.1001/jamacardio.2016.5670%5Cnhttp://www.ncbi.nlm.nih.gov/pubmed/28241245}.
  \DOIprefix\doi{10.1001/jamacardio.2016.5670}.
%Type = Inproceedings
\bibitem[{Ioffe and Szegedy(2015)}]{Ioffe2015}
\bibinfo{author}{Ioffe\xfnm[ S.]}, \bibinfo{author}{Szegedy\xfnm[ C.]}.
\newblock \bibinfo{title}{{Batch Normalization: Accelerating Deep Network
  Training by Reducing Internal Covariate Shift}}.
\newblock In: \bibinfo{booktitle}{ICML}. volume~\bibinfo{volume}{37};
  \bibinfo{year}{2015}. p. \bibinfo{pages}{81--87}.
\newblock \URLprefix \url{http://arxiv.org/abs/1502.03167}.
  \DOIprefix\doi{10.1007/s13398-014-0173-7.2}.
  \href{http://arxiv.org/abs/arXiv:1011.1669v3}{\tt arXiv:arXiv:1011.1669v3}.
%Type = Inproceedings
\bibitem[{Isensee et~al.(2017)Isensee, Jaeger, Full, Wolf, Engelhardt and
  Maier-Hein}]{Isensee2017}
\bibinfo{author}{Isensee\xfnm[ F.]}, \bibinfo{author}{Jaeger\xfnm[ P.]},
  \bibinfo{author}{Full\xfnm[ P.M.]}, \bibinfo{author}{Wolf\xfnm[ I.]},
  \bibinfo{author}{Engelhardt\xfnm[ S.]}, \bibinfo{author}{Maier-Hein\xfnm[
  K.H.]}.
\newblock \bibinfo{title}{{Automatic Cardiac Disease Assessment on cine-MRI via
  Time-Series Segmentation and Domain Specific Features}}.
\newblock In: \bibinfo{booktitle}{STACOM}. \bibinfo{address}{Quebec City,
  Quebec, Canada}; \bibinfo{year}{2017}. .
%Type = Unpublished
\bibitem[{Isensee et~al.(2018)Isensee, Jaeger, Full, Wolf, Engelhardt and
  Maier-Hein}]{Isensee2018}
\bibinfo{author}{Isensee\xfnm[ F.]}, \bibinfo{author}{Jaeger\xfnm[ P.]},
  \bibinfo{author}{Full\xfnm[ P.M.]}, \bibinfo{author}{Wolf\xfnm[ I.]},
  \bibinfo{author}{Engelhardt\xfnm[ S.]}, \bibinfo{author}{Maier-Hein\xfnm[
  K.H.]}.
\newblock \bibinfo{title}{{Automatic Cardiac Disease Assessment on cine-MRI via
  Time-Series Segmentation and Domain Specific Features}};
  \bibinfo{year}{2018}.
\newblock \URLprefix \url{http://arxiv.org/abs/1707.00587}.
  \href{http://arxiv.org/abs/1707.00587}{\tt arXiv:1707.00587}.
%Type = Inproceedings
\bibitem[{Jaderberg et~al.(2015)Jaderberg, Simonyan, Zisserman and
  Kavukcuoglu}]{Jaderberg2015}
\bibinfo{author}{Jaderberg\xfnm[ M.]}, \bibinfo{author}{Simonyan\xfnm[ K.]},
  \bibinfo{author}{Zisserman\xfnm[ A.]}, \bibinfo{author}{Kavukcuoglu\xfnm[
  K.]}.
\newblock \bibinfo{title}{{Spatial Transformer Networks}}.
\newblock In: \bibinfo{booktitle}{NIPS}. \bibinfo{year}{2015}. p.
  \bibinfo{pages}{1--14}.
\newblock \DOIprefix\doi{10.1038/nbt.3343}.
  \href{http://arxiv.org/abs/arXiv:1506.02025v1}{\tt arXiv:arXiv:1506.02025v1}.
%Type = Inproceedings
\bibitem[{Kingma and Ba(2015)}]{Kingma2015}
\bibinfo{author}{Kingma\xfnm[ D.]}, \bibinfo{author}{Ba\xfnm[ J.]}.
\newblock \bibinfo{title}{{Adam: A Method for Stochastic Optimization}}.
\newblock In: \bibinfo{booktitle}{ICLR}. \bibinfo{year}{2015}.
  \href{http://arxiv.org/abs/arXiv:1412.6980v9}{\tt arXiv:arXiv:1412.6980v9}.
%Type = Inproceedings
\bibitem[{Krizhevsky et~al.(2012)Krizhevsky, Sutskever and
  Hinton}]{Krizhevsky2012}
\bibinfo{author}{Krizhevsky\xfnm[ A.]}, \bibinfo{author}{Sutskever\xfnm[ I.]},
  \bibinfo{author}{Hinton\xfnm[ G.E.]}.
\newblock \bibinfo{title}{{ImageNet Classification with Deep Convolutional
  Neural Networks}}.
\newblock In: \bibinfo{booktitle}{NIPS}. \bibinfo{year}{2012}. .
%Type = Inproceedings
\bibitem[{Lieman-Sifry et~al.(2017)Lieman-Sifry, Le, Lau, Sall and
  Golden}]{Lieman-Sifry2017}
\bibinfo{author}{Lieman-Sifry\xfnm[ J.]}, \bibinfo{author}{Le\xfnm[ M.]},
  \bibinfo{author}{Lau\xfnm[ F.]}, \bibinfo{author}{Sall\xfnm[ S.]},
  \bibinfo{author}{Golden\xfnm[ D.]}.
\newblock \bibinfo{title}{{FastVentricle: Cardiac Segmentation with ENet}}.
\newblock In: \bibinfo{booktitle}{FIMH}. \bibinfo{year}{2017}. \URLprefix
  \url{http://arxiv.org/abs/1704.04296}.
  \href{http://arxiv.org/abs/1704.04296}{\tt arXiv:1704.04296}.
%Type = Inproceedings
\bibitem[{Long et~al.(2015)Long, Shelhamer and Darrell}]{Long2015}
\bibinfo{author}{Long\xfnm[ J.]}, \bibinfo{author}{Shelhamer\xfnm[ E.]},
  \bibinfo{author}{Darrell\xfnm[ T.]}.
\newblock \bibinfo{title}{{Fully convolutional networks for semantic
  segmentation}}.
\newblock In: \bibinfo{booktitle}{CVPR}. volume \bibinfo{volume}{07-12-June};
  \bibinfo{year}{2015}. p. \bibinfo{pages}{3431--3440}.
\newblock \DOIprefix\doi{10.1109/CVPR.2015.7298965}.
  \href{http://arxiv.org/abs/1411.4038}{\tt arXiv:1411.4038}.
%Type = Article
\bibitem[{Luo et~al.(2016)Luo, An, Wang, Dong and Zhang}]{Luo2016}
\bibinfo{author}{Luo\xfnm[ G.]}, \bibinfo{author}{An\xfnm[ R.]},
  \bibinfo{author}{Wang\xfnm[ K.]}, \bibinfo{author}{Dong\xfnm[ S.]},
  \bibinfo{author}{Zhang\xfnm[ H.]}.
\newblock \bibinfo{title}{{A Deep Learning Network for Right Ventricle
  Segmentation in Short-Axis MRI}}.
\newblock \bibinfo{journal}{Computing in Cardiology}
  \bibinfo{year}{2016};\bibinfo{volume}{43}:\bibinfo{pages}{485--488}.
%Type = Inproceedings
\bibitem[{Newell et~al.(2016)Newell, Yang and Deng}]{Newell2016}
\bibinfo{author}{Newell\xfnm[ A.]}, \bibinfo{author}{Yang\xfnm[ K.]},
  \bibinfo{author}{Deng\xfnm[ J.]}.
\newblock \bibinfo{title}{{Stacked Hourglass Networks for Human Pose
  Estimation}}.
\newblock In: \bibinfo{booktitle}{ECCV}. \bibinfo{year}{2016}. \URLprefix
  \url{http://arxiv.org/abs/1603.06937}.
  \DOIprefix\doi{10.1007/978-3-319-46484-8}.
  \href{http://arxiv.org/abs/1603.06937}{\tt arXiv:1603.06937}.
%Type = Inproceedings
\bibitem[{Noh et~al.(2015)Noh, Hong and Han}]{Noh2015}
\bibinfo{author}{Noh\xfnm[ H.]}, \bibinfo{author}{Hong\xfnm[ S.]},
  \bibinfo{author}{Han\xfnm[ B.]}.
\newblock \bibinfo{title}{{Learning Deconvolution Network for Semantic
  Segmentation}}.
\newblock In: \bibinfo{booktitle}{ICCV}. volume~\bibinfo{volume}{1};
  \bibinfo{year}{2015}. \URLprefix \url{http://arxiv.org/abs/1505.04366}.
  \DOIprefix\doi{10.1109/ICCV.2015.178}.
  \href{http://arxiv.org/abs/1505.04366}{\tt arXiv:1505.04366}.
%Type = Misc
\bibitem[{Paszke et~al.(2016)Paszke, Chaurasia, Kim and
  Culurciello}]{Paszke2016}
\bibinfo{author}{Paszke\xfnm[ A.]}, \bibinfo{author}{Chaurasia\xfnm[ A.]},
  \bibinfo{author}{Kim\xfnm[ S.]}, \bibinfo{author}{Culurciello\xfnm[ E.]}.
\newblock \bibinfo{title}{{ENet: A Deep Neural Network Architecture for
  Real-Time Semantic Segmentation}}.
\newblock \bibinfo{year}{2016}.
\newblock \URLprefix \url{https://arxiv.org/abs/1606.02147}.
  \href{http://arxiv.org/abs/arXiv:1606.02147v1}{\tt arXiv:arXiv:1606.02147v1}.
%Type = Article
\bibitem[{Peng et~al.(2016)Peng, Lekadir, Gooya, Shao, Petersen and
  Frangi}]{Peng2016}
\bibinfo{author}{Peng\xfnm[ P.]}, \bibinfo{author}{Lekadir\xfnm[ K.]},
  \bibinfo{author}{Gooya\xfnm[ A.]}, \bibinfo{author}{Shao\xfnm[ L.]},
  \bibinfo{author}{Petersen\xfnm[ S.E.]}, \bibinfo{author}{Frangi\xfnm[ A.F.]}.
\newblock \bibinfo{title}{{A review of heart chamber segmentation for
  structural and functional analysis using cardiac magnetic resonance
  imaging}}.
\newblock \bibinfo{journal}{MAGMA}
  \bibinfo{year}{2016};\bibinfo{volume}{29}(\bibinfo{number}{2}):\bibinfo{pages}{155--195}.
\newblock \DOIprefix\doi{10.1007/s10334-015-0521-4}.
%Type = Inproceedings
\bibitem[{Poudel et~al.(2016)Poudel, Lamata and Montana}]{Poudel2016a}
\bibinfo{author}{Poudel\xfnm[ R.P.]}, \bibinfo{author}{Lamata\xfnm[ P.]},
  \bibinfo{author}{Montana\xfnm[ G.]}.
\newblock \bibinfo{title}{{Recurrent Fully Convolutional Neural Networks for
  Multi-slice MRI Cardiac Segmentation}}.
\newblock In: \bibinfo{booktitle}{HVSMR}. \bibinfo{year}{2016}. \URLprefix
  \url{http://arxiv.org/abs/1608.03974}.
  \DOIprefix\doi{10.1007/978-3-319-52280-7_8}.
  \href{http://arxiv.org/abs/1608.03974}{\tt arXiv:1608.03974}.
%Type = Inproceedings
\bibitem[{Ronneberger et~al.(2015)Ronneberger, Fischer and
  Brox}]{Ronneberger2015}
\bibinfo{author}{Ronneberger\xfnm[ O.]}, \bibinfo{author}{Fischer\xfnm[ P.]},
  \bibinfo{author}{Brox\xfnm[ T.]}.
\newblock \bibinfo{title}{{U-Net: Convolutional Networks for Biomedical Image
  Segmentation}}.
\newblock In: \bibinfo{booktitle}{MICCAI}. \bibinfo{year}{2015}. p.
  \bibinfo{pages}{234--241}.
\newblock \DOIprefix\doi{10.1007/978-3-319-24574-4_28}.
  \href{http://arxiv.org/abs/1505.04597}{\tt arXiv:1505.04597}.
%Type = Inproceedings
\bibitem[{Saxe et~al.(2014)Saxe, McClelland and Ganguli}]{Saxe2013}
\bibinfo{author}{Saxe\xfnm[ A.M.]}, \bibinfo{author}{McClelland\xfnm[ J.L.]},
  \bibinfo{author}{Ganguli\xfnm[ S.]}.
\newblock \bibinfo{title}{{Exact solutions to the nonlinear dynamics of
  learning in deep linear neural networks}}.
\newblock In: \bibinfo{booktitle}{ICLR}. \bibinfo{year}{2014}. p.
  \bibinfo{pages}{1--22}.
\newblock \URLprefix \url{http://arxiv.org/abs/1312.6120}.
  \href{http://arxiv.org/abs/1312.6120}{\tt arXiv:1312.6120}.
%Type = Article
\bibitem[{Sifre and Mallat(2013)}]{Sifre2013}
\bibinfo{author}{Sifre\xfnm[ L.]}, \bibinfo{author}{Mallat\xfnm[ S.]}.
\newblock \bibinfo{title}{{Rotation, scaling and deformation invariant
  scattering for texture discrimination}}.
\newblock \bibinfo{journal}{Proceedings of the IEEE Computer Society Conference
  on Computer Vision and Pattern Recognition}
  \bibinfo{year}{2013};:\bibinfo{pages}{1233--1240}\DOIprefix\doi{10.1109/CVPR.2013.163}.
%Type = Inproceedings
\bibitem[{Simonyan and Zisserman(2014)}]{Simonyan2015}
\bibinfo{author}{Simonyan\xfnm[ K.]}, \bibinfo{author}{Zisserman\xfnm[ A.]}.
\newblock \bibinfo{title}{{Very Deep Convolutional Networks for Large-Scale
  Image Recognition}}.
\newblock In: \bibinfo{booktitle}{ICLR}. \bibinfo{year}{2014}. p.
  \bibinfo{pages}{1--14}.
\newblock \href{http://arxiv.org/abs/arXiv:1409.1556v6}{\tt
  arXiv:arXiv:1409.1556v6}.
%Type = Inproceedings
\bibitem[{Sinclair et~al.(2017)Sinclair, Bai, Puyol-Ant{\'{o}}n, Oktay,
  Rueckert and King}]{Sinclair2017}
\bibinfo{author}{Sinclair\xfnm[ M.]}, \bibinfo{author}{Bai\xfnm[ W.]},
  \bibinfo{author}{Puyol-Ant{\'{o}}n\xfnm[ E.]}, \bibinfo{author}{Oktay\xfnm[
  O.]}, \bibinfo{author}{Rueckert\xfnm[ D.]}, \bibinfo{author}{King\xfnm[
  A.P.]}.
\newblock \bibinfo{title}{{Fully Automated Segmentation-Based Respiratory
  Motion Correction of Multiplanar Cardiac Magnetic Resonance Images for
  Large-Scale Datasets}}.
\newblock In: \bibinfo{booktitle}{MICCAI}. volume~\bibinfo{volume}{2};
  \bibinfo{year}{2017}. p. \bibinfo{pages}{332--340}.
\newblock \URLprefix
  \url{http://link.springer.com/10.1007/978-3-319-66185-8_38}.
  \DOIprefix\doi{10.1007/978-3-319-66185-8_38}.
%Type = Inproceedings
\bibitem[{Tan et~al.(2016)Tan, Liew, Lim and Mclaughlin}]{Tan2016}
\bibinfo{author}{Tan\xfnm[ L.K.]}, \bibinfo{author}{Liew\xfnm[ Y.M.]},
  \bibinfo{author}{Lim\xfnm[ E.]}, \bibinfo{author}{Mclaughlin\xfnm[ R.A.]}.
\newblock \bibinfo{title}{{Cardiac Left Ventricle Segmentation using
  Convolutional Neural Network Regression}}.
\newblock In: \bibinfo{booktitle}{IECBES}. \bibinfo{year}{2016}. p.
  \bibinfo{pages}{490--493}.
%Type = Article
\bibitem[{Tan et~al.(2017)Tan, Liew, Lim and McLaughlin}]{Tan2017}
\bibinfo{author}{Tan\xfnm[ L.K.]}, \bibinfo{author}{Liew\xfnm[ Y.M.]},
  \bibinfo{author}{Lim\xfnm[ E.]}, \bibinfo{author}{McLaughlin\xfnm[ R.A.]}.
\newblock \bibinfo{title}{{Convolutional neural network regression for
  short-axis left ventricle segmentation in cardiac cine MR sequences}}.
\newblock \bibinfo{journal}{Medical Image Analysis}
  \bibinfo{year}{2017};\bibinfo{volume}{39}:\bibinfo{pages}{78--86}.
\newblock \URLprefix
  \url{https://www.sciencedirect.com/science/article/pii/S1361841517300543}.
  \DOIprefix\doi{10.1016/j.media.2017.04.002}.
%Type = Unpublished
\bibitem[{Tran(2016)}]{Tran2016}
\bibinfo{author}{Tran\xfnm[ P.V.]}.
\newblock \bibinfo{title}{{A Fully Convolutional Neural Network for Cardiac
  Segmentation in Short-Axis MRI}}; \bibinfo{year}{2016}.
\newblock \URLprefix \url{http://arxiv.org/abs/1604.00494}.
  \href{http://arxiv.org/abs/1604.00494}{\tt arXiv:1604.00494}.
%Type = Book
\bibitem[{Vigneault et~al.(2017)Vigneault, Xie, Bluemke and
  Noble}]{Vigneault2017}
\bibinfo{author}{Vigneault\xfnm[ D.]}, \bibinfo{author}{Xie\xfnm[ W.]},
  \bibinfo{author}{Bluemke\xfnm[ D.]}, \bibinfo{author}{Noble\xfnm[ J.]}.
\newblock \bibinfo{title}{{Feature tracking cardiac magnetic resonance via deep
  learning and spline optimization}}.
\newblock volume \bibinfo{volume}{10263 LNCS}, \bibinfo{year}{2017}.
\newblock \DOIprefix\doi{10.1007/978-3-319-59448-4_18}.
%Type = Inproceedings
\bibitem[{Viola and Jones(2001)}]{Viola2001}
\bibinfo{author}{Viola\xfnm[ P.]}, \bibinfo{author}{Jones\xfnm[ M.]}.
\newblock \bibinfo{title}{{Rapid Object Detection using a Boosted Cascade of
  Simple Features}}.
\newblock In: \bibinfo{booktitle}{CVPR}. \bibinfo{year}{2001}. .
%Type = Inproceedings
\bibitem[{Xie et~al.(2015)Xie, Noble and Zisserman}]{Xie2015}
\bibinfo{author}{Xie\xfnm[ W.]}, \bibinfo{author}{Noble\xfnm[ J.A.]},
  \bibinfo{author}{Zisserman\xfnm[ A.]}.
\newblock \bibinfo{title}{{Microscopy Cell Counting with Fully Convolutional
  Regression Networks}}.
\newblock In: \bibinfo{booktitle}{MICCAI Workshop}. \bibinfo{year}{2015}. p.
  \bibinfo{pages}{1--10}.
\newblock \URLprefix
  \url{http://www.tandfonline.com.myaccess.library.utoronto.ca/doi/full/10.1080/21681163.2016.1149104}.
  \DOIprefix\doi{10.1080/21681163.2016.1149104}.
%Type = Inproceedings
\bibitem[{Yu and Koltun(2016)}]{Yu2016}
\bibinfo{author}{Yu\xfnm[ F.]}, \bibinfo{author}{Koltun\xfnm[ V.]}.
\newblock \bibinfo{title}{{Multi-Scale Context Aggregation by Dilated
  Convolutions}}.
\newblock In: \bibinfo{booktitle}{ICLR}. \bibinfo{year}{2016}. p.
  \bibinfo{pages}{1--9}.
\newblock \URLprefix \url{http://arxiv.org/abs/1511.07122}.
  \DOIprefix\doi{10.16373/j.cnki.ahr.150049}.
  \href{http://arxiv.org/abs/1511.07122}{\tt arXiv:1511.07122}.
%Type = Article
\bibitem[{Yushkevich et~al.(2006)Yushkevich, Piven, Hazlett, Smith, Ho, Gee and
  Gerig}]{Yushkevich2006}
\bibinfo{author}{Yushkevich\xfnm[ P.A.]}, \bibinfo{author}{Piven\xfnm[ J.]},
  \bibinfo{author}{Hazlett\xfnm[ H.C.]}, \bibinfo{author}{Smith\xfnm[ R.G.]},
  \bibinfo{author}{Ho\xfnm[ S.]}, \bibinfo{author}{Gee\xfnm[ J.C.]},
  \bibinfo{author}{Gerig\xfnm[ G.]}.
\newblock \bibinfo{title}{{User-guided 3D active contour segmentation of
  anatomical structures : Significantly improved efficiency and reliability}}.
\newblock \bibinfo{journal}{NeuroImage}
  \bibinfo{year}{2006};\bibinfo{volume}{31}:\bibinfo{pages}{1116--1128}.
\newblock \DOIprefix\doi{10.1016/j.neuroimage.2006.01.015}.

\end{thebibliography}
